\documentclass[12pt,twoside]{article}
\usepackage[dvips]{graphics}
\graphicspath{{./Images/}}
\usepackage{epsfig}
\usepackage{amssymb}
\usepackage{amsmath}
\usepackage[utf8]{inputenc}

\usepackage[hypertex]{hyperref} 
\usepackage[british]{babel} 
\usepackage[thinlines]{easytable} 

\usepackage{float} 
\usepackage{listings}

\usepackage{color} 
\definecolor{blue}{rgb}{0.,0.,1.0}      
\definecolor{lightblue}{rgb}{0.4,0.5,0.8}      
\definecolor{mygreen}{rgb}{0,0.4,0}

\usepackage{my}
\def\i{$\infty$}  
\def\setR{\mathrm I\!\mathbf{R}}  

\begin{document}
\pagestyle{myheadings}
\markboth{T.Strutz: The Distance Transform and its Computation}{TECHP/2021/06, v2/2023/02}
\title{\vspace{-3em}\bf The Distance Transform and its Computation\\--- An Introduction  ---}
\author{\copyright Tilo Strutz\\%
 Technical paper, 2021, June, TECHP/2021/06\\%
update: \today \\%
 Leipzig University of Telecommunication}
\date{\vspace{-5ex}} 
\maketitle
\begin{abstract}
Distance transformation is an image processing technique used for many different applications. Related to a binary image, the general idea is to determine the distance of all background points to the nearest object point (or vice versa). In this tutorial, different approaches are explained in detail and compared using examples. Corresponding source code is provided to facilitate own investigations. A particular objective of this tutorial is to clarify the difference between arbitrary distance transforms and {\bf exact Euclidean} distance transformations.
\end{abstract}
   %
\section{General description}
\label{sec_allgemein}
    %
\subsection{Examples and possible applications}
\begin{figure}
	\hfil (a)\includegraphics[height=5cm]{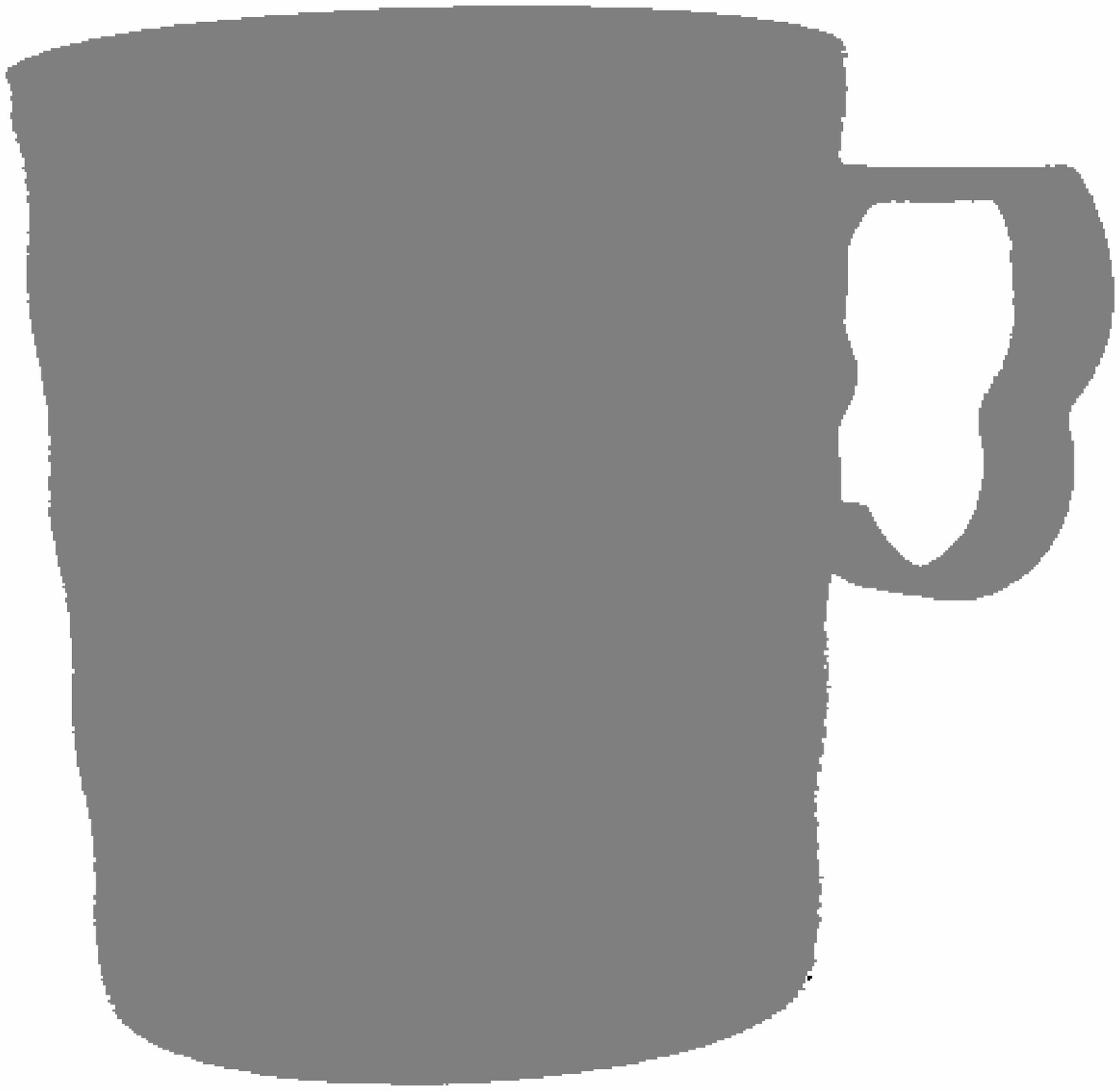}
	\hfil (b)\includegraphics[height=5cm]{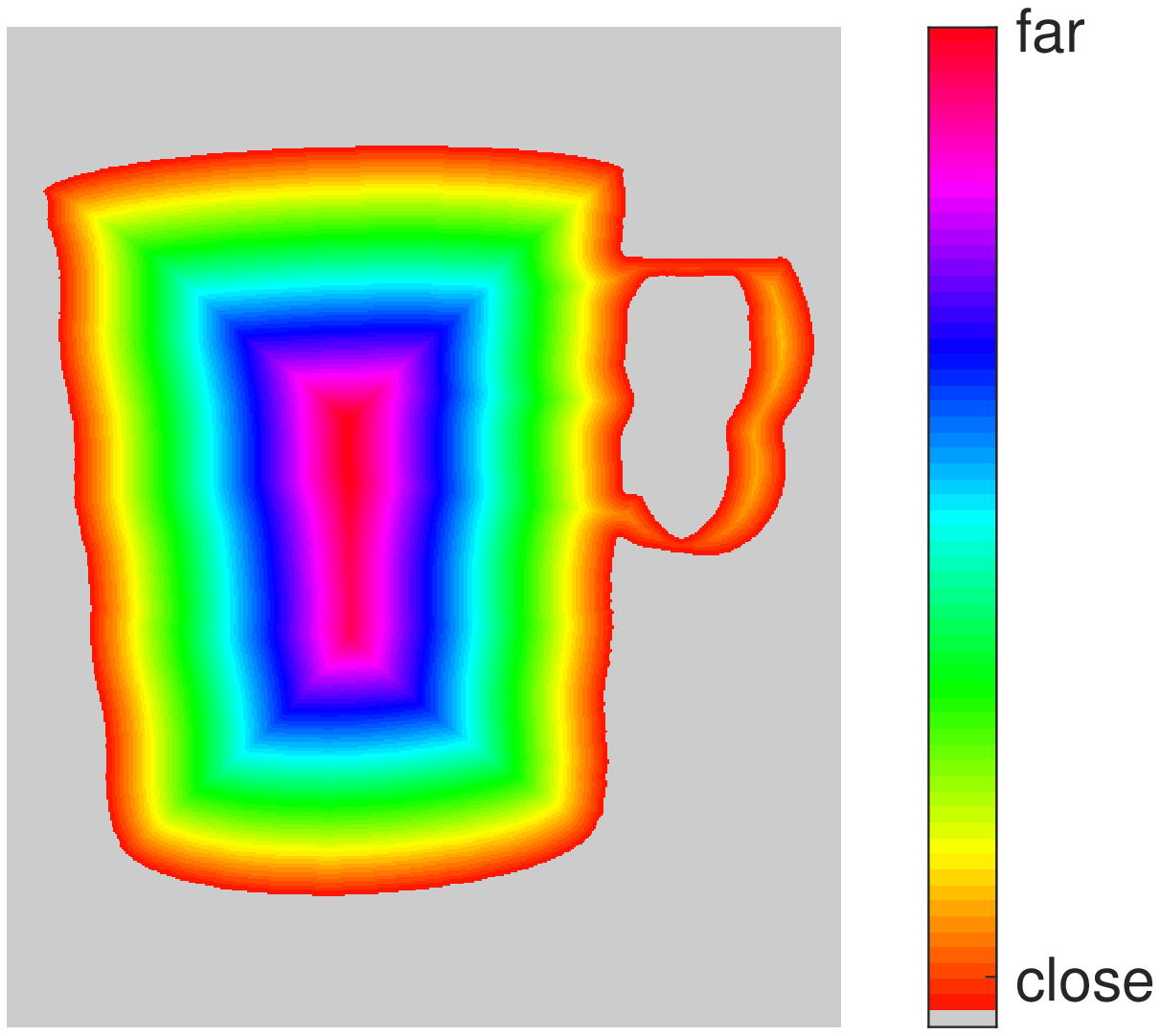}
	\hfil (c)\includegraphics[height=5cm]{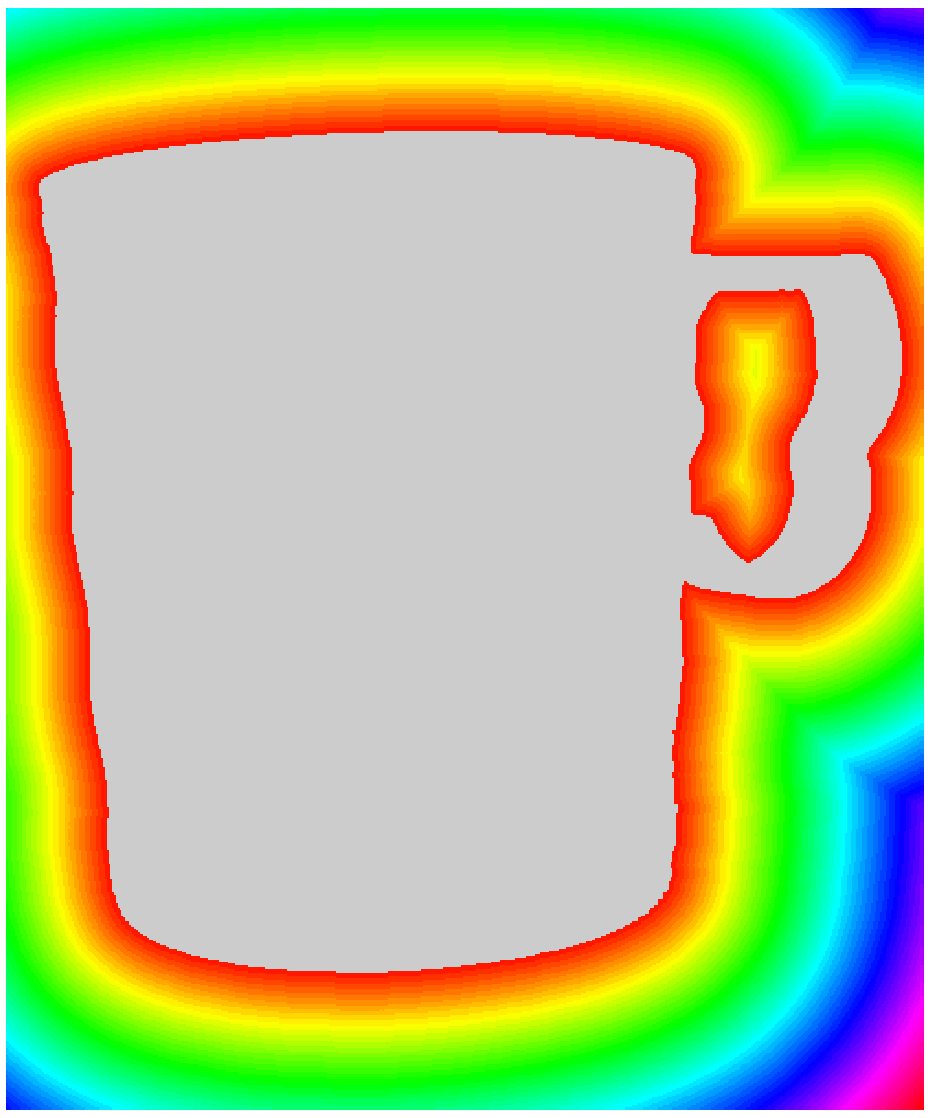}
	\caption{\label{fig_examples}Example of a distance transform: a) binary image containing a dark object on white background; (b)+(c) results of distance transform showing the distance of object pixels to closest background pixel (b) and the distance of background pixels to closest object pixel (c)}
\end{figure}

\Figu{fig_examples} shows what can be achieved in general by a distance transform.
The input is typically a binary image (Figure \ref{fig_examples}a), i. e. the pixels can have only one out of two different values. One value is associated to the background and the other value defines the object pixels. The object pixels can belong to one or more (disconnected) objects. Depending on the application, either the distances inside the objects are of interest (Figure \ref{fig_examples}b) or the distances outside the objects (Figure \ref{fig_examples}c).

\begin{figure}
	\hfil (a) \includegraphics[scale=0.4]{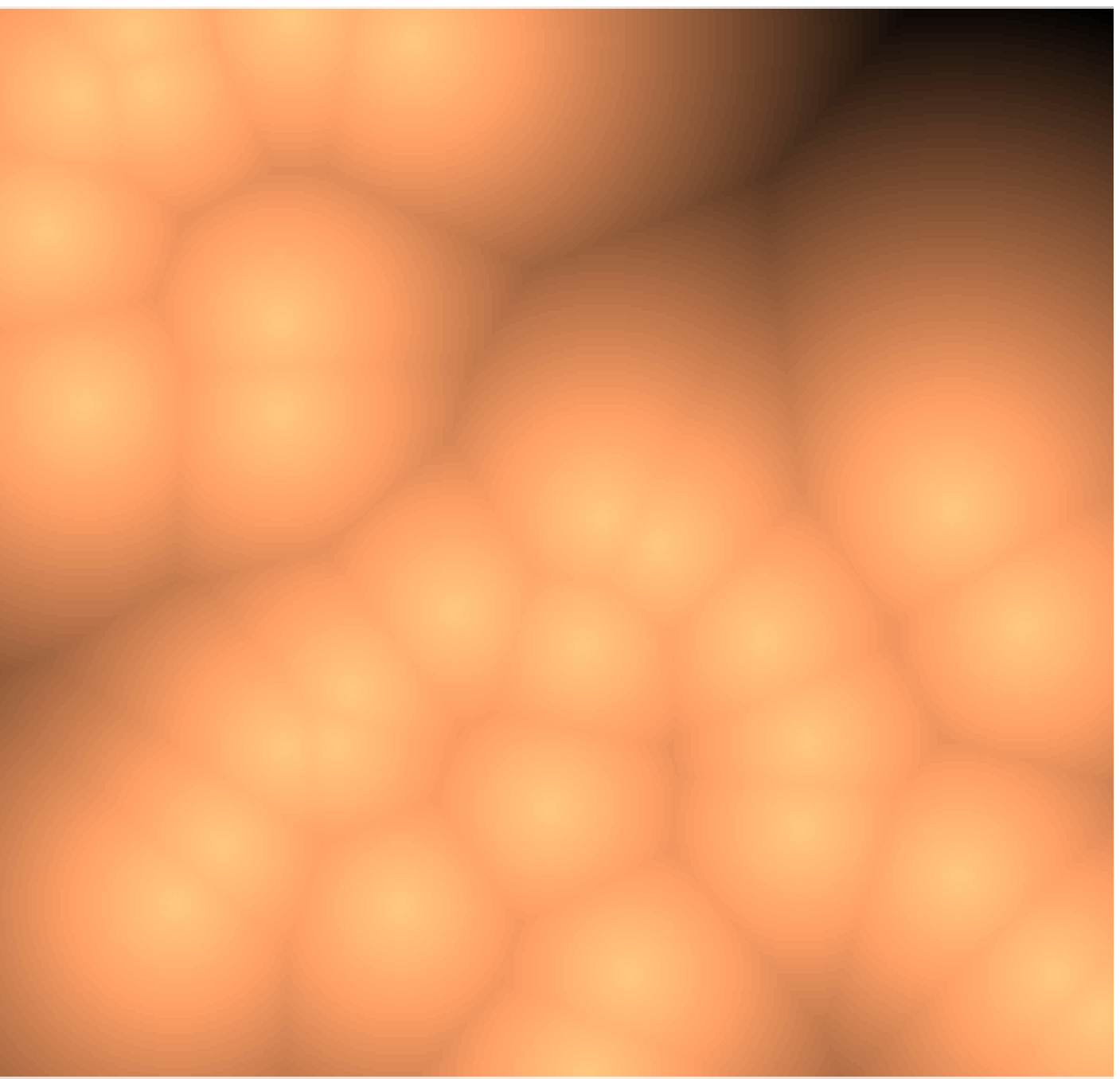}
	\hfil (b) \includegraphics[scale=0.4]{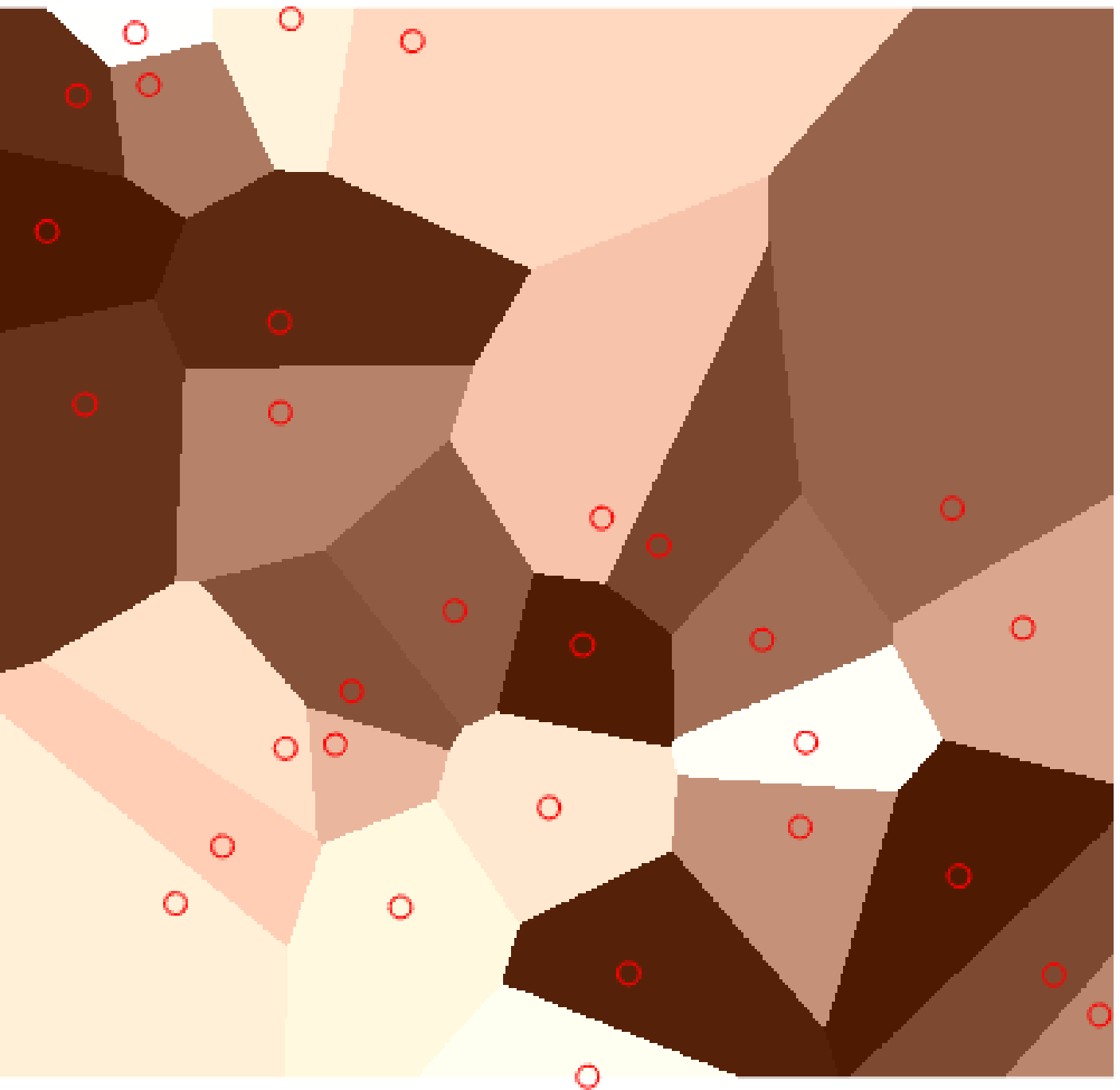}
	\caption{\label{fig_voronoi}Example image with 30 separate object points: (a) distance matrix; (b) Voronoi diagram with marked positions of object points.}
\end{figure}

\begin{figure}[tp]
	(a) \includegraphics[scale=0.23]{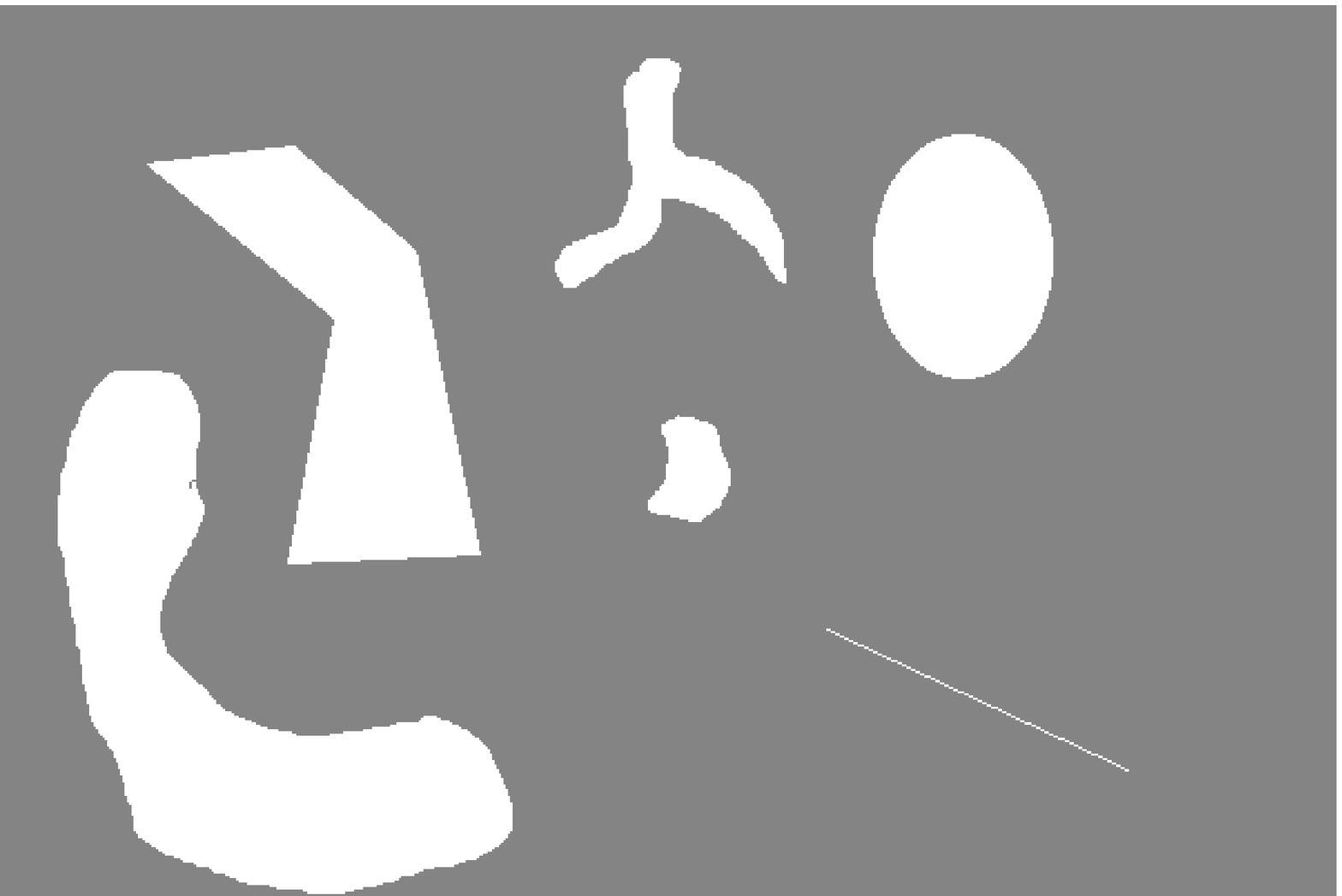}
	 (b) \includegraphics[scale=0.35]{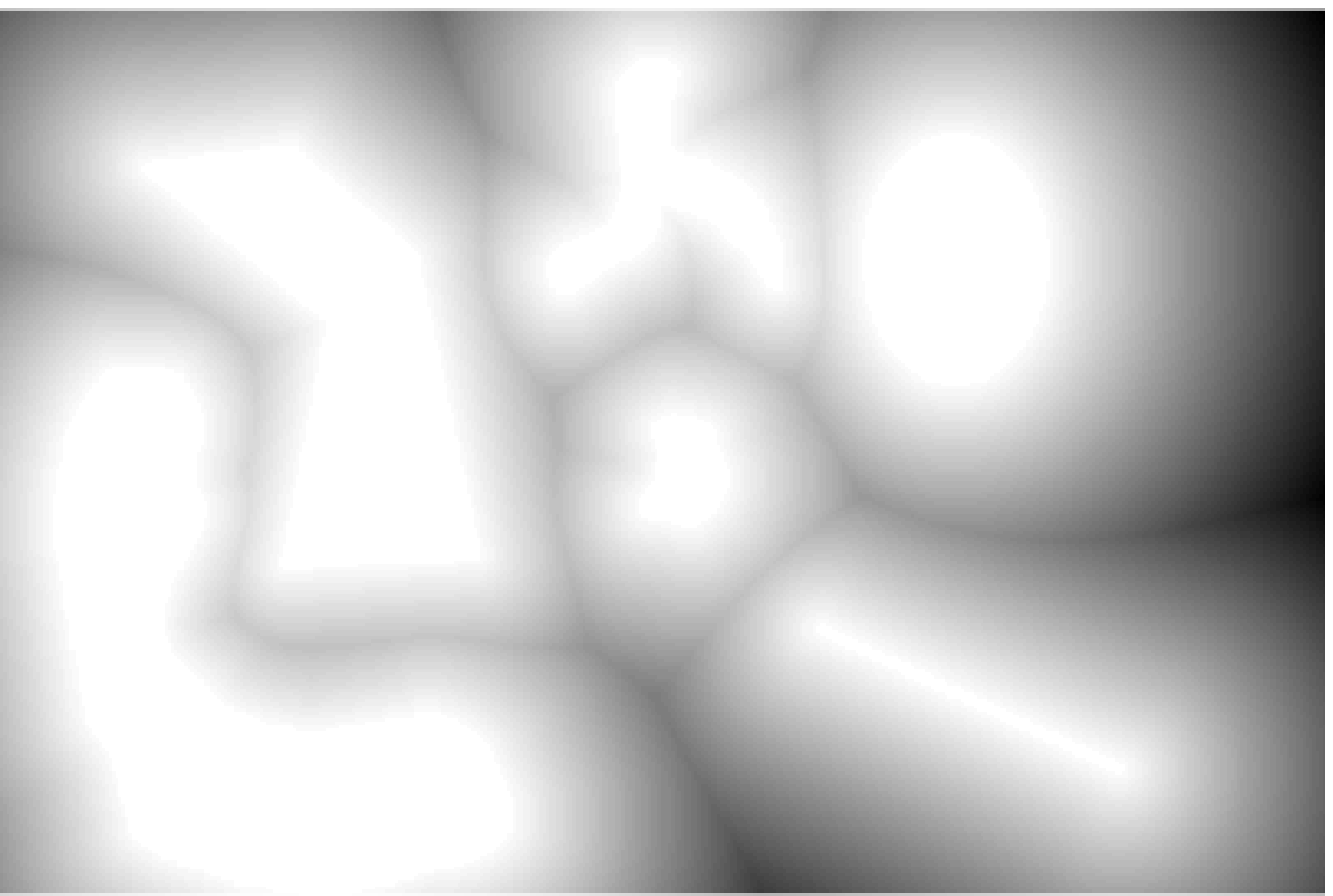}
	 (c) \includegraphics[scale=0.35]{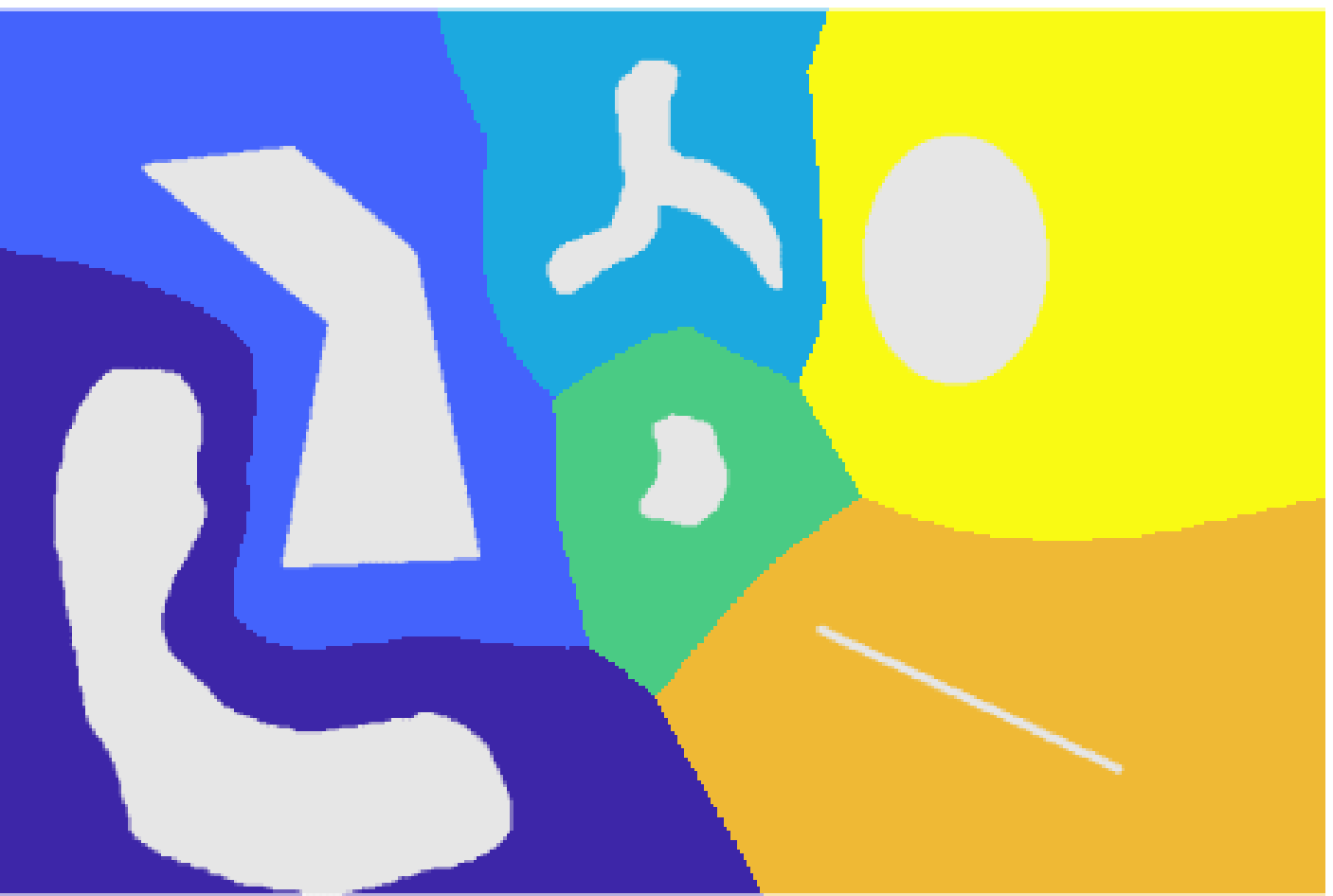}
	\caption{\label{fig_blobs}Example image containing six binary objects: (a) original binary image; (b) distance matrix; (c) Voronoi diagram.}
\end{figure}

The segmentation of images into regions is one important application of the distance transformation. Afterwards, each region represents one object point or a cluster of connected points. This is highly related to so-called Voronoi diagram. \Figu{fig_voronoi} and \Figu{fig_blobs} shows two examples.
From these Voronoi diagrams it can be deduced which points are adjacent. This information could be helpful in solving graph-based problems.	

\begin{figure}
	\hfil (a) \includegraphics[scale=0.6]{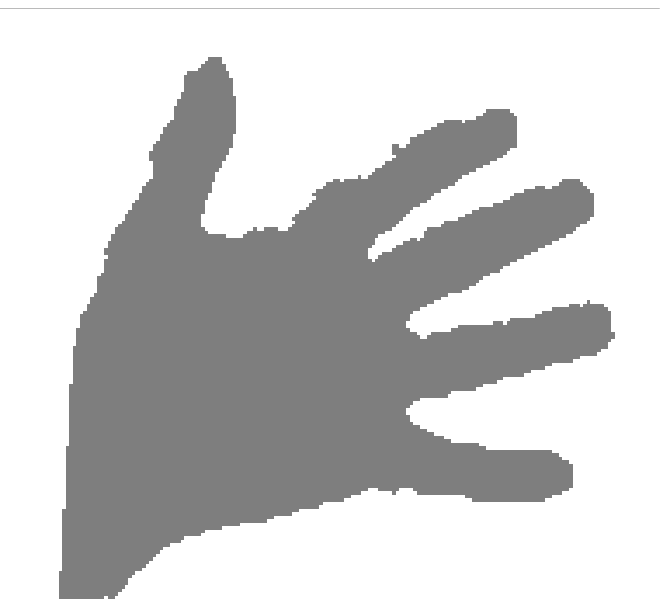}
	\hfil (b) \includegraphics[scale=0.33]{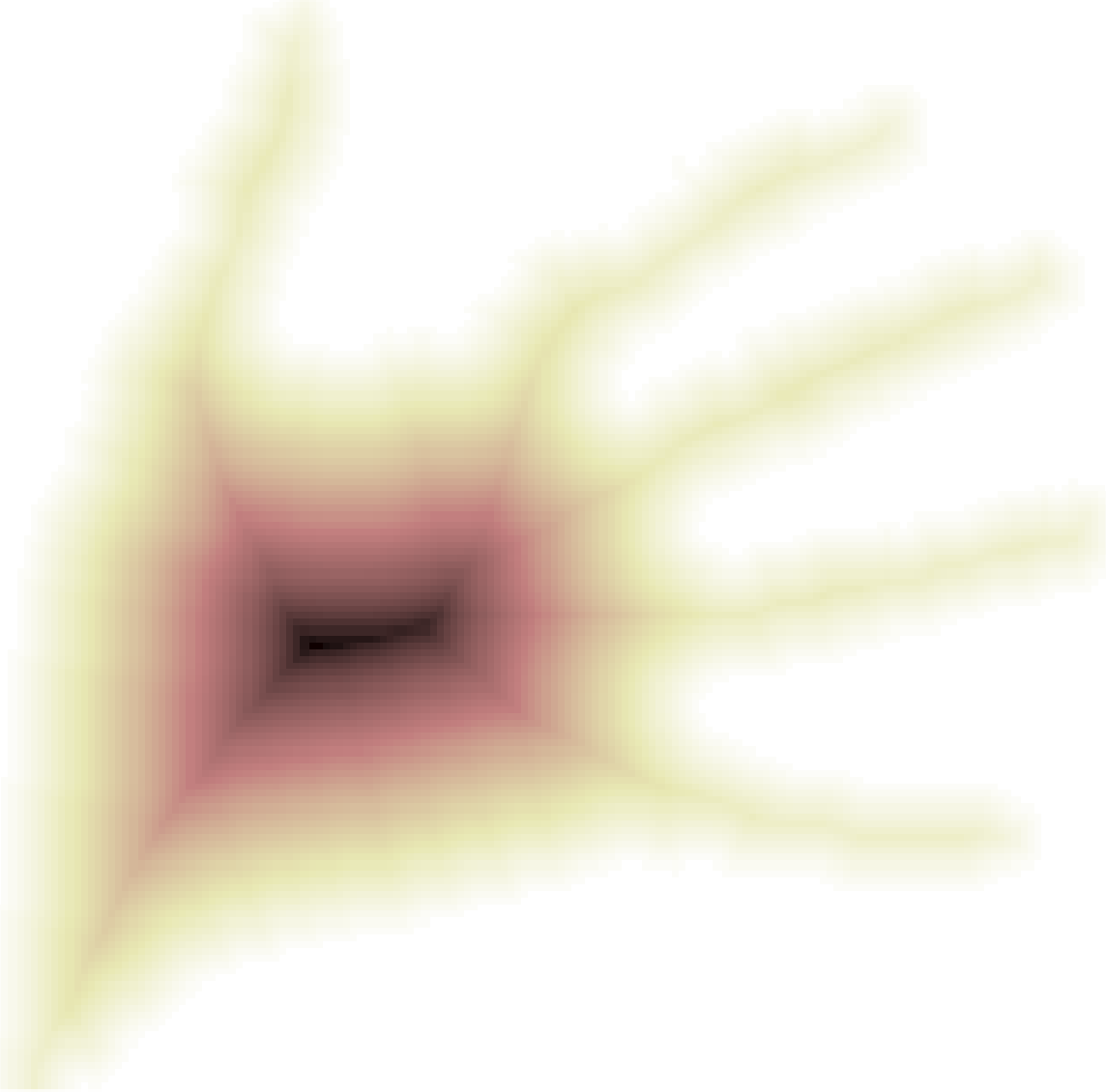}
	\hfil (c) \includegraphics[scale=0.6]{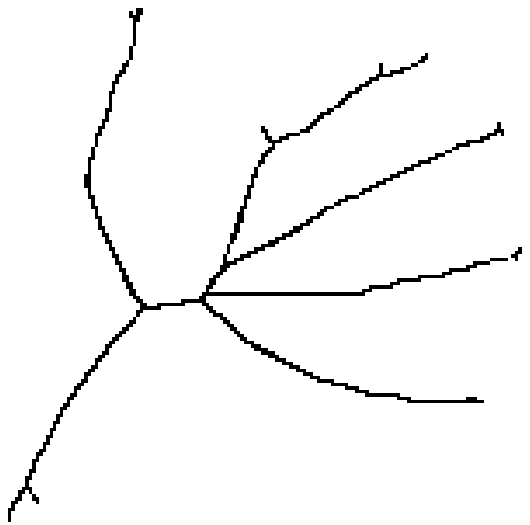}
	\caption{\label{fig_hand}Steps in hand-gesture recognition: (a) original binary image; (b) distance matrix for finding the centre of the palm; (c) derived skeleton.}
\end{figure}

\begin{figure}
	\hfil (a) \includegraphics[scale=0.22]{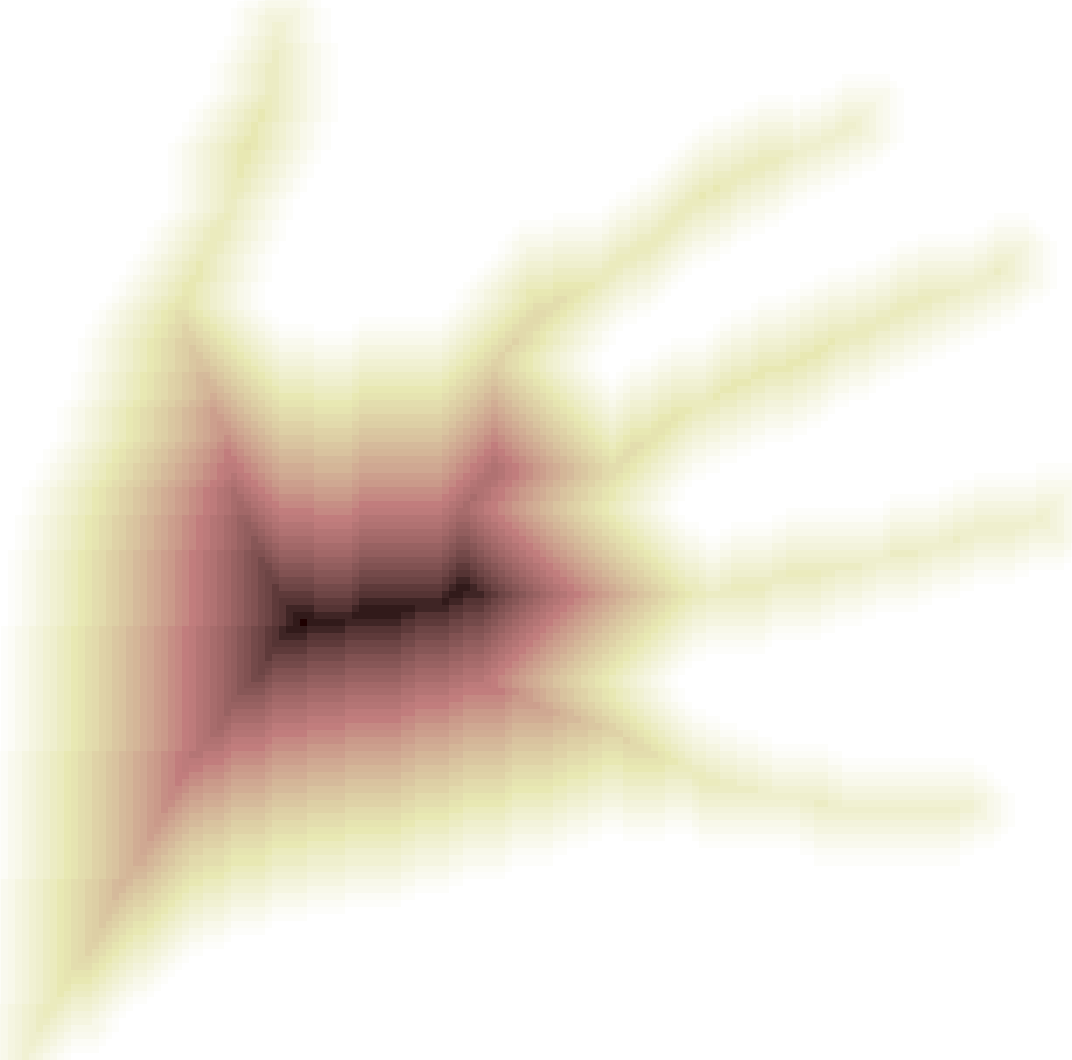}
	\hfil (b) \includegraphics[scale=0.4]{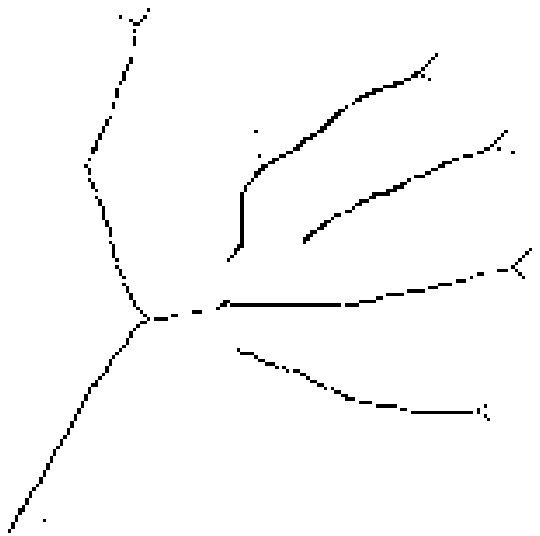}
	\hfil (c) \includegraphics[scale=0.22]{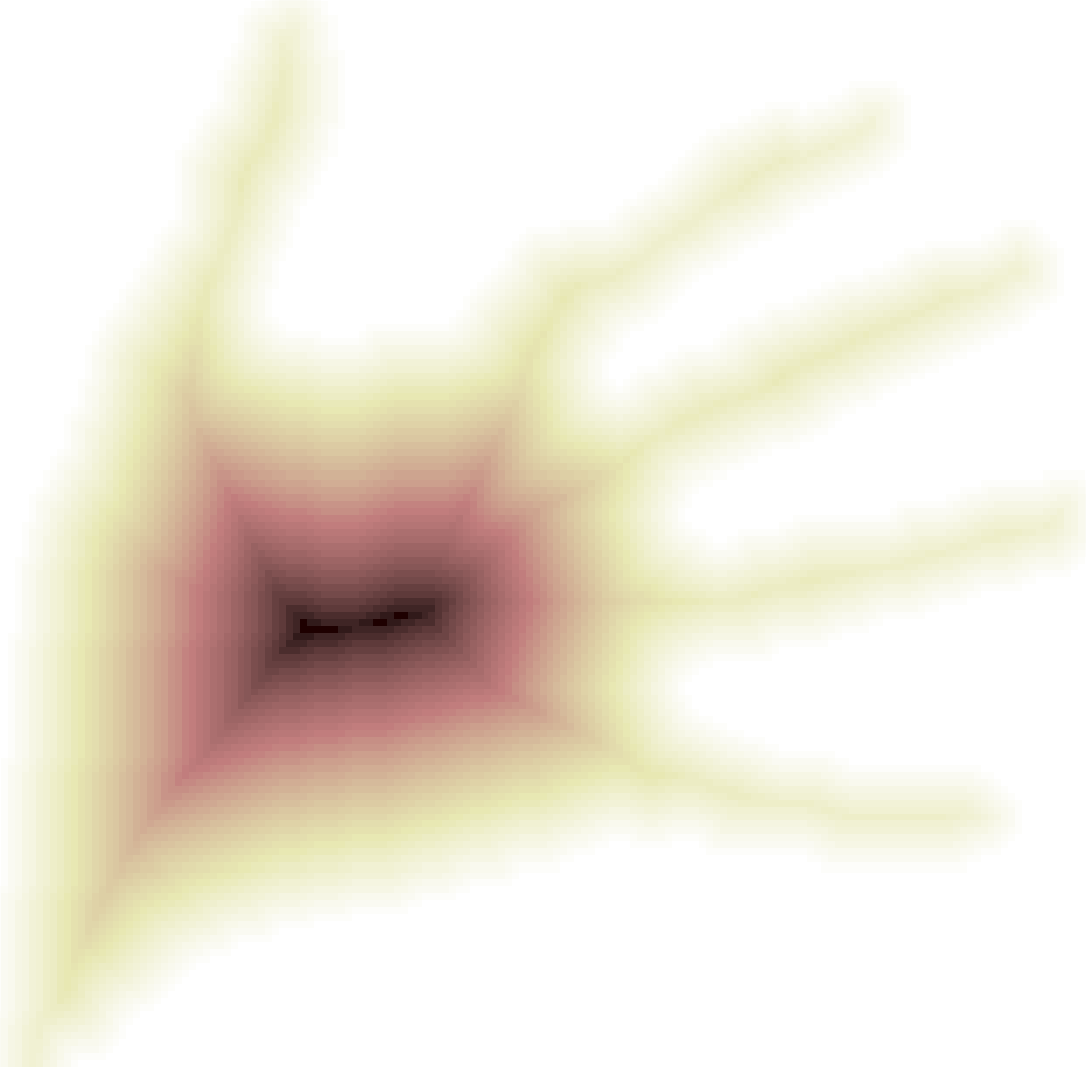}
	\hfil (d) \includegraphics[scale=0.4]{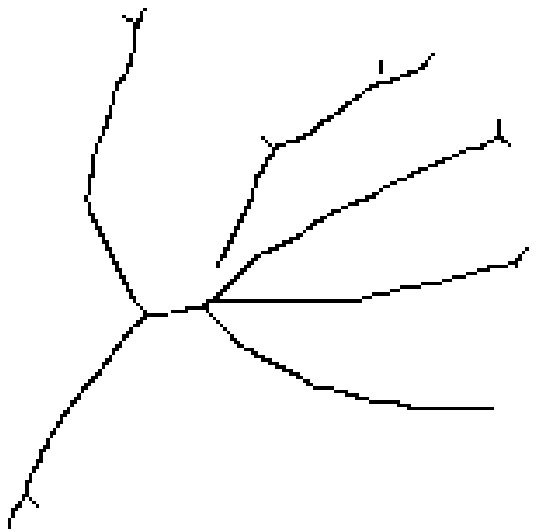}
	\caption{\label{fig_hand2}Results for non-Euclidean distances: (a) distance matrix based on city-block distance; (b) derived skeleton; (c) distance matrix based on chamfer-43 distance; (d) derived skeleton.}
\end{figure}

\begin{figure}
	\hfil  \includegraphics[scale=0.25]{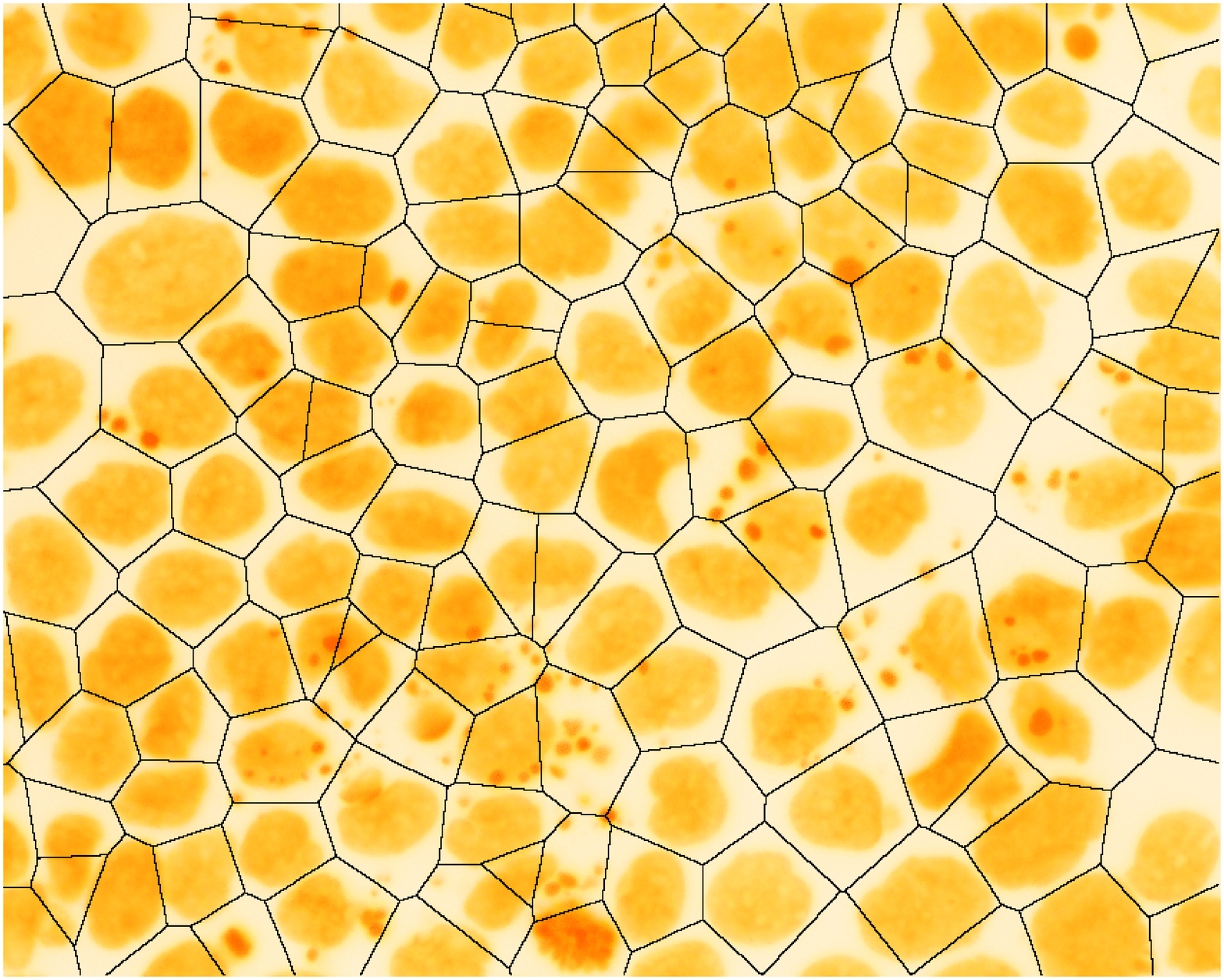}
	\caption{\label{fig_cellSeparation}Separation of objects utilizing  distance transforms.}
\end{figure}

The distance within objects also could be of interest for finding the centre of objects like hands or for the skeletonization of objects, see \Figu{fig_hand}.
Intermediate processing results may vary when different distance metrics (which are discussed in Subsection \ref{subsec_metrics}) are used, see \Figu{fig_hand2}.
And last but not least, distance transforms can support the segmentation of overlapping objects like cells, \Figu{fig_cellSeparation}.

This paper wont discuss, how the distance transform has to be used in achieving the goals of these applications; instead, it concentrates on the basics of distances and on algorithms for computing a distance transformation in an appropriate manner.
First ideas about distances in digital images have been discussed in a pioneering work by Rosenfeld and Pfaltz in 1968 \cite{Ros68}.

The following two Subsections discuss neighbourhood relations and different metrics for distances. If you are already familiar with it, just skip to Section \ref{sec_approximateMethods}.
	%
\subsection{Neighbourhood relations}
\begin{figure}
	\hfil (a)\includegraphics[scale=0.6]{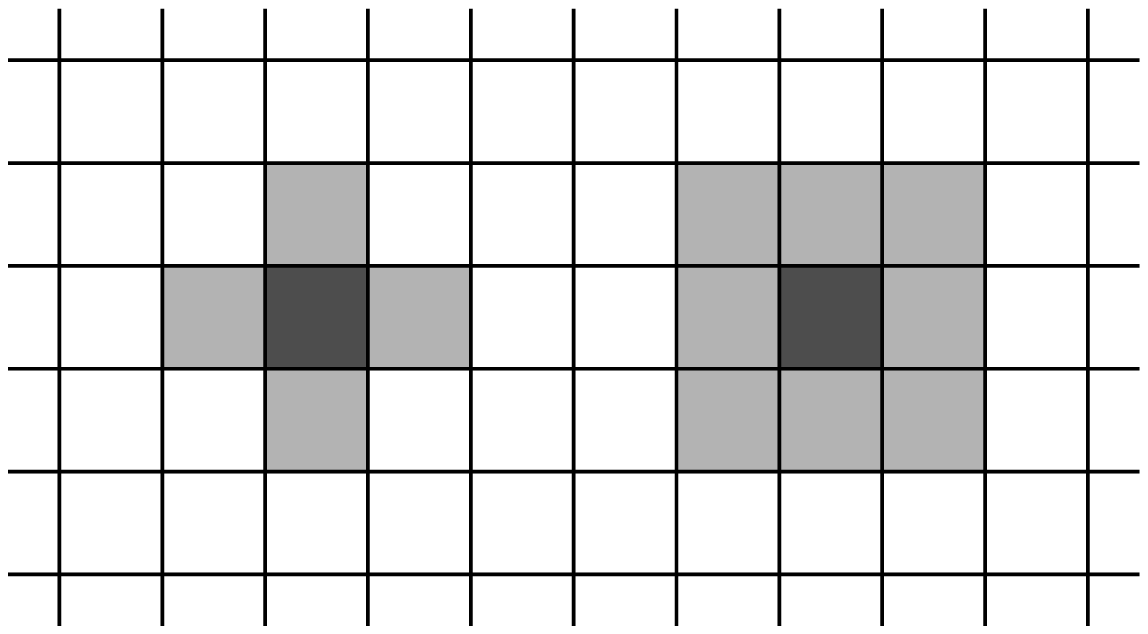}
	\hfil (b)\includegraphics[scale=0.6]{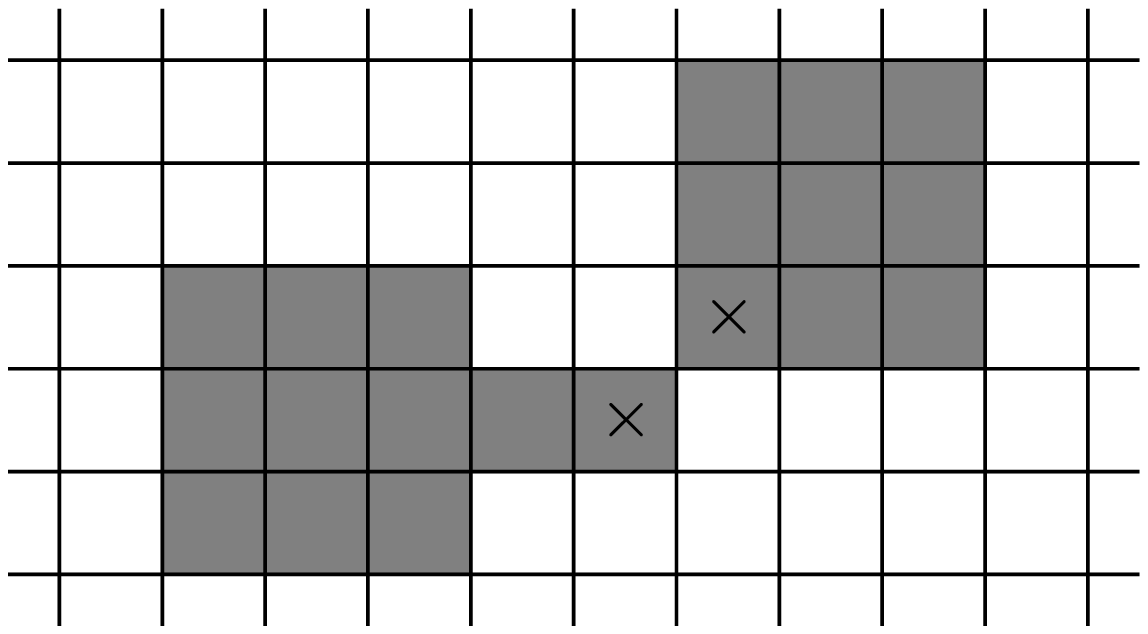}
	\caption{\label{fig_neighbourhoods}Neighbourhood configurations: (a) 4-neighbourhood (left) and 8-neigh\-bour\-hood (right); (b) dark object pixels are considered as a single object in 8-neighbourhood but as two (not connected) objects in 4-neighbourhood.}
\end{figure}

In an orthogonal grid of pixels, we can differentiate between two neighbourhood relations. The first is called 4-neighbourhood since only the horizontal and vertical neighbouring pixels are considered. When also the diagonal pixels are taken into account it is called 8-neighbourhood, \Figu{fig_neighbourhoods}(a).

This differentiation has implications in several directions. If, for example, binary image objects have to be counted, the result can be different. \Figu{fig_neighbourhoods}(b) shows dark object pixels, which belong to a single object. In 8-neighbourhood, all pixels belong to a single object, because the marked pixels are considered as neighbours. However, in 4-neighbourhood the marked pixels are not connected and there are two binary objects.
	%
	%
\subsection{Distance Metrics}\label{subsec_metrics}
	%
\subsubsection{Minkowski metric}
\begin{figure}
	\hfil (a)\includegraphics[scale=0.6]{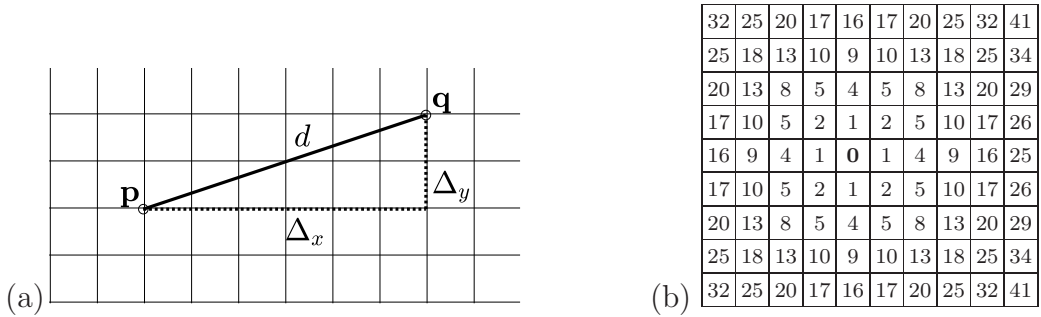}
	\hfil (b)
	{\scriptsize
\begin{TAB}(e,0.44cm,0.44cm){1c1c1c1c1c1c1c1c1c1c1}{1c1c1c1c1c1c1c1c1c1}
 32 & 25 	& 20	& 17 & 16 	& 17 &20 & 25 & 32 & 41 \\
 25 & 18 	& 13 	& 10 & 9 		& 10 &13 & 18 & 25 & 34 \\
 20 & 13 	& 8		& 5  & 4 		& 5  & 8 & 13 & 20 & 29 \\
 17 & 10 	& 5 	& 2  & 1 		& 2  & 5 & 10 & 17 & 26 \\
 16 & 9 	& 4 	& 1  &{\bf 0}&1  & 4 & 9  & 16 & 25 \\
 17 & 10 	& 5 	& 2  & 1    & 2  & 5 & 10 & 17 & 26 \\
 20 & 13	& 8 	& 5  & 4 		& 5  & 8 & 13 & 20 & 29 \\
 25 & 18 	& 13 	& 10 & 9 		& 10 &13 & 18 & 25 & 34 \\
 32 & 25 	& 20	& 17 & 16 	& 17 &20 & 25 & 32 & 41 
\end{TAB}
}
	\caption{\label{fig_Euclidean}Euclidean distance in two dimensions: (a) $\Delta_x = q_x-p_x$, $\Delta_y = q_y-p_y$, the distance is $d= \sqrt{\Delta_x^2 + \Delta_y^2}$; (b) calculated squared distances $d^2$ from a centre pixel.}
\end{figure}

The distances shown in Figure \ref{fig_examples} are based on the Euclidean metric. This is what we also know as beeline distance. If we hammer two nails into a board and connect them with a rubber band, then the Euclidean distance is equal to the one-way length of the rubber band. In two dimensions this distance is defined by the Pythagorean theorem: $c^2 = a^2+b^2$.
If there are two points $\mathbf{p}$ and $\mathbf{q}$ with two-dimensional coordinates $(p_x, p_y)$ and $(q_x, q_y)$, then the Euclidean distance $d$ is computed with $d=\sqrt{ (p_x-q_x)^2 + (p_y-q_y)^2}$, see \Figu{fig_Euclidean}.

\begin{figure}
	\hfil (a)\includegraphics[scale=0.6]{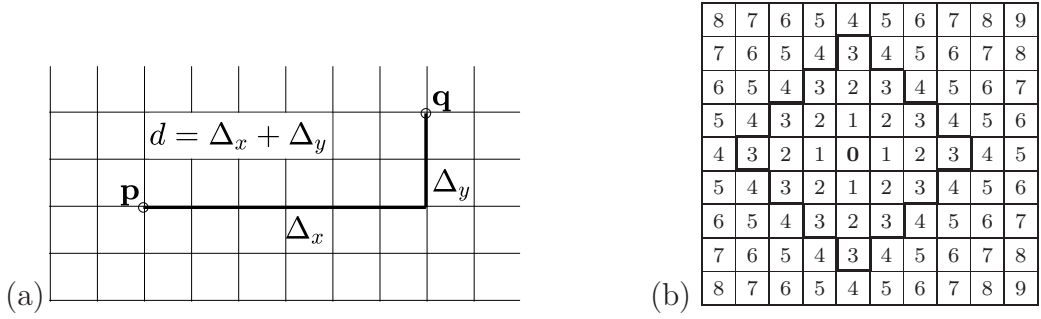}
	\hfil (b)
	{\scriptsize
\begin{TAB}(e,0.44cm,0.44cm){1c1c1c1c1c1c1c1c1c1c1}{1c1c1c1c1c1c1c1c1c1}
 8 & 7 	& 6	& 5 & 4 		& 5 &6 & 7 & 8 & 9 \\
 7 & 6 	& 5 & 4 & 3 		& 4 &5 & 6 & 7 & 8 \\
 6 & 5 	& 4	& 3 & 2 		& 3 &4 & 5 & 6 & 7 \\
 5 & 4 	& 3 & 2 & 1 		& 2 &3 & 4 & 5 & 6 \\
 4 & 3 	& 2 & 1 &{\bf 0}& 1 &2 & 3 & 4 & 5 \\
 5 & 4 	& 3 & 2 & 1    	& 2 &3 & 4 & 5 & 6 \\
 6 & 5	& 4 & 3 & 2 		& 3 &4 & 5 & 6 & 7 \\
 7 & 6 	& 5 & 4 & 3 		& 4 &5 & 6 & 7 & 8 \\
 8 & 7 	& 6	& 5 & 4 		& 5 &6 & 7 & 8 & 9 
\addpath{(4,1,4)rurururululululdldldldrdrdrd} 
\end{TAB}
}	
	\caption{\label{fig_cityblock}City-block distance in two dimensions: (a) $\Delta_x = q_x-p_x$, $\Delta_y = q_y-p_y$, distance with $d= \Delta_x + \Delta_y$; (b) calculated distances from a centre pixel.}
\end{figure}

Depending on the application, different metrics could be useful.
A more general definition is given by the Minkowski distance:
	\begin{align}\label{eq_Minkowski}
	  d(\mathbf{p},\mathbf{q}) = \left[ \sum_{i=1}^D |p_i - q_i|^e \right]^{1/e}
	\end{align}
with $D$ being the number of dimensions.
The Euclidean metric is a special case with $e=2$. Alternatively, $e=1$ corresponds to the city-block distance, see \Figu{fig_cityblock}.
This distance is sometimes called the Manhattan distance because most streets in Manhattan are either parallel or orthogonal to each other. If somebody wants to walk from point $\mathbf{p}$ to point $\mathbf{q}$, then she has to go along the streets and there is no possibility to cross the buildings.

\begin{figure}
	\hfil (a)\includegraphics[scale=0.6]{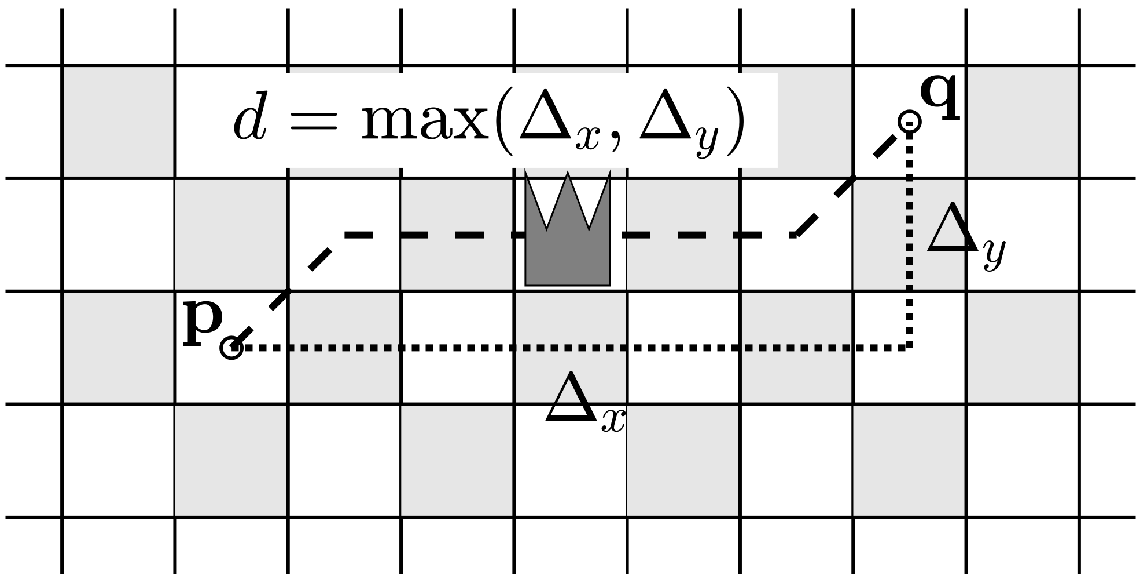}
	\hfil (b)
	{\scriptsize
\begin{TAB}(e,0.44cm,0.44cm){1c1c1c1c1c1c1c1c1c1c1}{1c1c1c1c1c1c1c1c1c1}
 4 & 4 	& 4	& 4 & 4 		& 4 &4 & 4 & 4 & 5 \\
 4 & 3 	& 3 & 3 & 3 		& 3 &3 & 3 & 4 & 5 \\
 4 & 3 	& 2	& 2 & 2 		& 2 &2 & 3 & 4 & 5 \\
 4 & 3 	& 2 & 1 & 1 		& 1 &2 & 3 & 4 & 5 \\
 4 & 3 	& 2 & 1 &{\bf 0}& 1 &2 & 3 & 4 & 5 \\
 4 & 3 	& 2 & 1 & 1    	& 1 &2 & 3 & 4 & 5 \\
 4 & 3	& 2 & 2 & 2 		& 2 &2 & 3 & 4 & 5 \\
 4 & 3 	& 3 & 3 & 3 		& 3 &3 & 3 & 4 & 5 \\
 4 & 4 	& 4	& 4 & 4 		& 4 &4 & 4 & 4 & 5 
\addpath{(4,1,4)rrrruuuuuuullllllldddddddrrr} 
\end{TAB}
}	\caption{\label{fig_chessboard}Chessboard distance in two dimensions $\Delta_x = q_x-p_x$, $\Delta_y = q_y-p_y$: (a) distance with $d= \max(\Delta_x, \Delta_y)$; (b) calculated distances from a centre pixel.}
\end{figure}

Setting variable $e$ to infinity defines the chessboard distance. This is the number of moves that the king has to take on the chessboard from one square to another, \Figu{fig_chessboard}.
	%

	%
\subsubsection{Chamfer metric}
Besides the metrics that can be derived from the Minkowski distance in (\ref{eq_Minkowski}), there are other ways to define distances. One of these is called `chamfer' distance and is constructed by using a different step size in diagonal directions compared to horizontal or vertical direction.
Both, the city-block and the chessboard distances apply steps of `1' in horizontal and vertical direction. However, the diagonal step is equal to `2' in the former and equal to `1' in the latter case as can be seen in Figures \ref{fig_cityblock}~(b) and \ref{fig_chessboard}~(b).

\begin{figure}
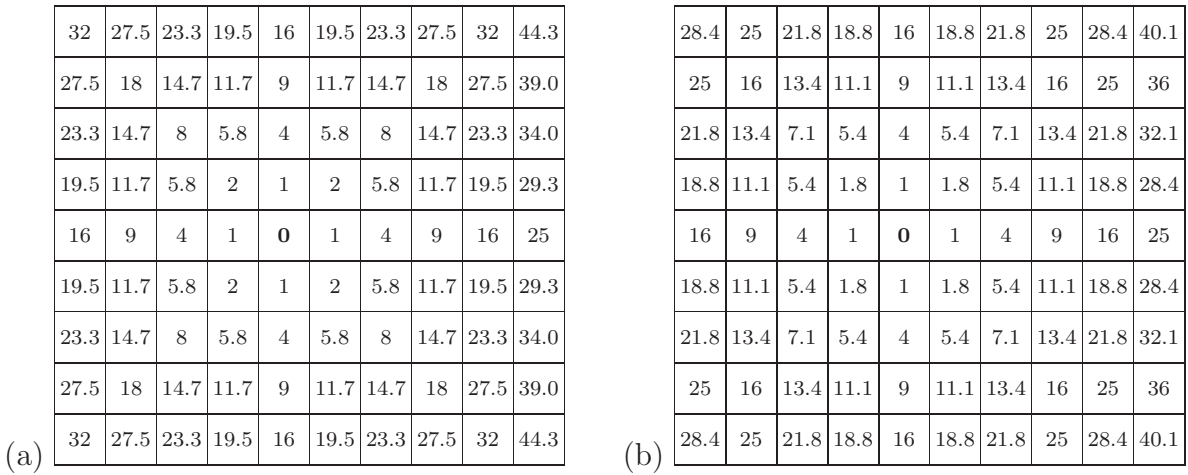

	\hfil (a)
	{\scriptsize
\begin{TAB}(e,0.44cm,0.44cm){1c1c1c1c1c1c1c1c1c1c1}{1c1c1c1c1c1c1c1c1c1}
 32	 & 27.5	&23.3 & 19.5 & 16 	& 19.5 &23.3 & 27.5 & 32	 & 44.3\\
 27.5& 18		&14.7 & 11.7 & 9 		& 11.7 &14.7 & 18		& 27.5 & 39.0 \\
 23.3& 14.7	& 8 	& 5.8  & 4 		& 5.8  & 8 	 & 14.7 & 23.3 & 34.0 \\
 19.5& 11.7	& 5.8 & 2		 & 1 		& 2		 & 5.8 & 11.7 & 19.5 & 29.3 \\
 16	 & 9 		& 4	 	& 1	   &{\bf 0}&1	   & 4	 & 9 		& 16	 & 25 \\
 19.5& 11.7	& 5.8 & 2 	 & 1    & 2 	 & 5.8 & 11.7 & 19.5 & 29.3 \\
 23.3& 14.7	& 8	 	& 5.8  & 4 		& 5.8  & 8	 & 14.7 & 23.3 & 34.0 \\
 27.5& 18		&14.7 & 11.7 & 9 		& 11.7 &14.7 & 18		& 27.5 & 39.0 \\
 32	 & 27.5	&23.3 & 19.5 & 16 	& 19.5 &23.3 & 27.5 & 32	 & 44.3 
\end{TAB}
}
	\hfil (b)
	{\scriptsize
\begin{TAB}(e,0.44cm,0.44cm){1c1c1c1c1c1c1c1c1c1c1}{1c1c1c1c1c1c1c1c1c1}
 28.4 &  25 	&21.8 & 18.8 & 16 	 & 18.8 &21.8 & 25 	& 28.4 & 40.1\\
 25	  &  16	  &13.4 & 11.1 & 9 		 & 11.1 &13.4 & 16	& 25	 & 36 \\
 21.8 &  13.4 & 7.1 & 5.4  & 4 		 & 5.4  & 7.1 & 13.4& 21.8 & 32.1 \\
 18.8 &  11.1 & 5.4 & 1.8	 & 1 		 & 1.8	& 5.4 & 11.1& 18.8 & 28.4 \\
 16	  &  9		& 4	 	& 1	   &{\bf 0}&1	  	& 4	 	& 9 	& 16	 & 25 \\
 18.8 &  11.1 & 5.4 & 1.8	 & 1     & 1.8	& 5.4 & 11.1& 18.8 & 28.4 \\
 21.8 &  13.4 & 7.1 & 5.4  & 4 		 & 5.4  & 7.1 & 13.4& 21.8 & 32.1 \\
 25 	&  16	  &13.4 & 11.1 & 9 		 & 11.1 &13.4 & 16	& 25 	 & 36 \\
 28.4 &  25 	&21.8 & 18.8 & 16 	 & 18.8 &21.8 & 25 	& 28.4 & 40.1 
\end{TAB}
}	
	\caption{\label{fig_chamfer}Calculated (and rounded) squared chamfer distances $d^2$ from a centre pixels. Exact values are shown as integers: 
	(a) with diagonal steps of $\sqrt{2}$; 
	(b) with diagonal steps of $4/3$.}
\end{figure}

As the Euclidean distance would fit many applications best, there have been made lots of attempts to approximate this metric by simpler constructions without the need of multiplications and drawing the square root.
So, one of the first ideas was to use a diagonal step of $\sqrt{2}$. The resulting effect can be seen in \Figu{fig_chamfer}(a).
The distance values directly close to the centre pixel, in horizontal, vertical, and diagonal direction are now exactly the same as for the Euclidean distance (Figure \ref{fig_Euclidean}b). At other positions, the distances differ. For example, the distance measure at the pixel in the bottom right corner deviates from the Euclidean distance by $\Delta d = \sqrt{44.3} -\sqrt{41}$. Borgefors has derived in \cite{Bor84} that the average and the maximum approximation error can be minimized by using a diagonal step of 1.351. 
That means, the optimal relation between diagonal and other steps is
	\begin{align}\label{eq_rel}
	  \frac{\D \mbox{step}_d}{\D \mbox{step}_{h,v}} = \frac{\D 1.351}{\D 1}
		\approx \frac{\D 1.\ol{3}}{\D 1} = \frac{\D 4}{\D 3}
	\end{align}
Using the approximation $4/3$, the distance computations can now be performed in integer arithmetic if all values are scaled by factor 3.
\Figu{fig_chamfer}~(b) shows the corresponding squared distances which are now closer to the squared Euclidean distances on average. The already mentioned pixel in the bottom-right corner has now a distance that is closer to the Euclidean one ($\left|\Delta d_{43}\right| =\left|\sqrt{40.1} -\sqrt{41}\right|  < \sqrt{44.3} -\sqrt{41}$), also compare with Figure \ref{fig_Euclidean}b.
However, the reduced diagonal step ($1.\ol{3}$ instead of $\sqrt{2}\approx 1.41$) leads to higher absolute differences at all positions that are directly diagonal to the centre pixel.

The values have been calculated using the code listed in \Listing{lst_distCalc} (see appendix).
	%
		%
\section{Non-Euclidean Distance Transforms}
\label{sec_approximateMethods}
 		%
This section explains several versions of distance transformations that can be implemented by fast algorithms but are not able to produce exact Euclidean distances.
If you are only interested in exact Euclidean distance transformations, then you can simply skip to the section \ref{sec_EEDT}.

In the following, the case shown in Figure \ref{fig_examples}~(c) is considered. That is, the shortest distances between a background pixel and any object pixel are searched for.
		%
\subsection{City-block distance transform}\label{subsec_cbDT}
One of the earliest ideas to perform a distance transformation was based on distance propagation. Starting from the positions of the object pixels, whose distances are zero by definition, the distances of the background pixels are determined step by step while incrementing the distance value.
This can be achieved by sequentially processing all image rows first from top-left to bottom-right and then in backward direction.

\begin{figure}
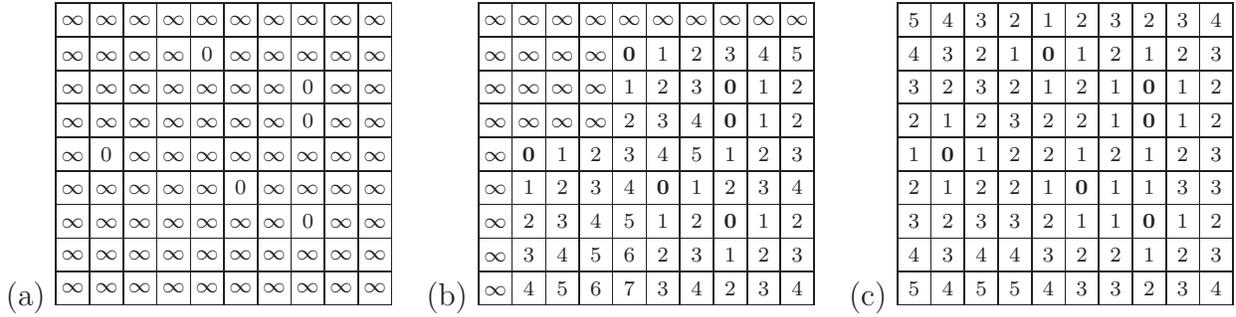

	\hfil (a)
	{\scriptsize
\begin{TAB}(e,0.44cm,0.44cm){1c1c1c1c1c1c1c1c1c1c1}{1c1c1c1c1c1c1c1c1c1}
 \i & \i 	& \i	& \i & \i & \i &\i & \i & \i & \i \\
 \i & \i 	& \i 	& \i & 0 	& \i &\i & \i & \i & \i \\
 \i & \i 	& \i	& \i & \i & \i &\i &  0 & \i & \i \\
 \i & \i 	& \i	& \i & \i & \i &\i &  0 & \i & \i \\
 \i & 0 	& \i	& \i & \i	& \i &\i & \i & \i & \i \\
 \i & \i 	& \i	& \i & \i &  0 &\i & \i & \i & \i \\
 \i & \i	& \i	& \i & \i & \i &\i &  0 & \i & \i \\
 \i & \i 	& \i 	& \i & \i & \i &\i & \i & \i & \i \\
 \i & \i 	& \i	& \i & \i & \i &\i & \i & \i & \i 
\end{TAB}
}
	\hfil (b)
	{\scriptsize
\begin{TAB}(e,0.44cm,0.44cm){1c1c1c1c1c1c1c1c1c1c1}{1c1c1c1c1c1c1c1c1c1}
 \i & \i 	& \i	& \i & \i		& \i 		&\i & \i 		& \i & \i \\
 \i & \i 	& \i 	& \i &{\bf 0}&  1 	& 2 &  3 		&  4 & 5 \\
 \i & \i 	& \i	& \i & 1		&   2 	& 3 &{\bf 0}&  1 &  2 \\
 \i & \i 	& \i	& \i & 2	 	&   3		& 4 &{\bf 0}&  1 &  2 \\
 \i &{\bf 0}& 1	&  2 & 3		&   4	 	& 5 &  1 		&  2 &  3 \\
 \i & 1 	&  2	&  3 & 4	 	&{\bf 0}& 1 &  2 		&  3 &  4 \\
 \i & 2		&  3	&  4 & 5	 	&  1 		& 2 &{\bf 0}&  1 &  2 \\
 \i & 3 	&  4 	&  5 & 6	 	&  2 		& 3 &  1 		&  2 &  3 \\
 \i & 4 	&  5	&  6 & 7	 	&  3 		& 4 &  2 		&  3 &  4 
\end{TAB}
}
	\hfil (c)
	{\scriptsize
\begin{TAB}(e,0.44cm,0.44cm){1c1c1c1c1c1c1c1c1c1c1}{1c1c1c1c1c1c1c1c1c1}
 5 &  4 	&  3	&  2 &  1		&   2 	& 3 &  2 		&  3 &  4 \\
 4 &  3 	&  2 	&  1 &{\bf 0}&  1 	& 2 &  1 		&  2 &  3 \\
 3 &  2 	&  3	&  2 & 1		&   2 	& 1 &{\bf 0}&  1 &  2 \\
 2 &  1 	&  2	&  3 & 2	 	&   2		& 1 &{\bf 0}&  1 &  2 \\
 1 &{\bf 0}& 1	&  2 & 2		&   1	 	& 2 &  1 		&  2 &  3 \\
 2 &  1 	&  2	&  2 & 1	 	&{\bf 0}& 1 &  1 		&  3 &  3 \\
 3 &  2		&  3	&  3 & 2	 	&  1 		& 1 &{\bf 0}&  1 &  2 \\
 4 &  3 	&  4 	&  4 & 3	 	&  2 		& 2 &  1 		&  2 &  3 \\
 5 &  4 	&  5	&  5 & 4	 	&  3 		& 3 &  2 		&  3 &  4 
\end{TAB}
}
	\caption{\label{fig_cityblockDT}Computation of sequential city-block distances: (a) initialisation with zero distances at object-pixel positions; (b) calculation of distances from the top-left down to the bottom-right; (c) final result}
\end{figure}

At first, the distance image (or matrix) has to be initialized with values of zero at the positions of object pixels and with sufficient large value (larger than the maximum possible distance) at all other positions. An example is given in \Figu{fig_cityblockDT}(a).
Then all rows are scanned from left to right. If the distance $d(i,j)$ value at the current position is larger than the predecessor plus one $d(i,j-1)+1$, it is set to a new value $d(i,j):=d(i,j-1)+1$. Afterwards, the same comparison is done with the position above:
	\begin{align*}
		\mbox{forward scan for all} (i,j)\left\{
		\begin{array}{lcl}
			\mbox{if} & d(i,j) > d(i,j-1) + 1 & d(i,j) := d(i,j-1)+1 \\
			\mbox{if} & d(i,j) > d(i-1,j) + 1 & d(i,j) := d(i-1,j)+1 \\
		\end{array}
		\right.
		\;.
	\end{align*}
This procedure propagates the distances to the right and to the bottom as shown in \Figu{fig_cityblockDT}(b). This direction of propagation has to be reversed in a second scan from the bottom-right up to the top-left position:
	\begin{align*}
		\mbox{backward scan for all} (i,j)\left\{
		\begin{array}{lcl}
			\mbox{if} & d(i,j) > d(i,j+1) + 1 & d(i,j) := d(i,j+1)+1 \\
			\mbox{if} & d(i,j) > d(i+1,j) + 1 & d(i,j) := d(i+1,j)+1 \\
		\end{array}
		\right.
		\;,
	\end{align*}
and the final result in \Figu{fig_cityblockDT}(c) is obtained.
The corresponding source code is shown in \Listing{lst_cityblockDTsequ} (see appendix).
In total, the image is scanned twice and two comparisons per scan have to be made at each pixel position.

\begin{figure}
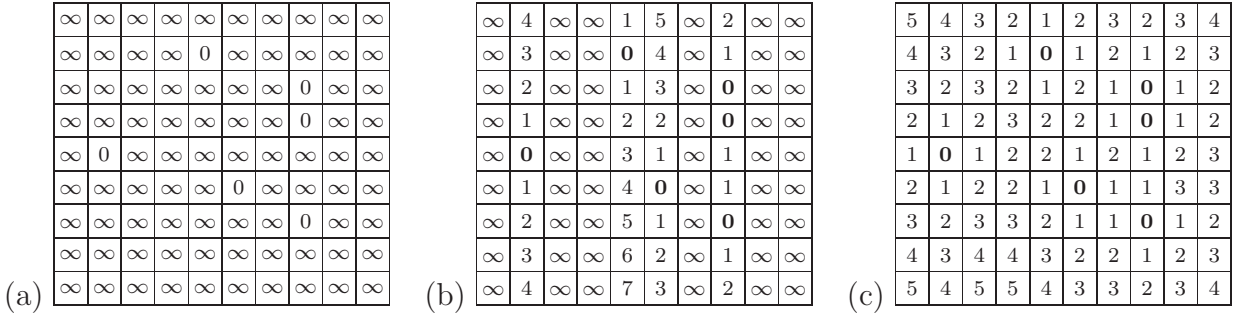

	\hfil (a)
	{\scriptsize
\begin{TAB}(e,0.44cm,0.44cm){1c1c1c1c1c1c1c1c1c1c1}{1c1c1c1c1c1c1c1c1c1}
 \i & \i 	& \i	& \i & \i & \i &\i & \i & \i & \i \\
 \i & \i 	& \i 	& \i & 0 	& \i &\i & \i & \i & \i \\
 \i & \i 	& \i	& \i & \i & \i &\i &  0 & \i & \i \\
 \i & \i 	& \i	& \i & \i & \i &\i &  0 & \i & \i \\
 \i & 0 	& \i	& \i & \i	& \i &\i & \i & \i & \i \\
 \i & \i 	& \i	& \i & \i &  0 &\i & \i & \i & \i \\
 \i & \i	& \i	& \i & \i & \i &\i &  0 & \i & \i \\
 \i & \i 	& \i 	& \i & \i & \i &\i & \i & \i & \i \\
 \i & \i 	& \i	& \i & \i & \i &\i & \i & \i & \i 
\end{TAB}
}
	\hfil (b)
	{\scriptsize
\begin{TAB}(e,0.44cm,0.44cm){1c1c1c1c1c1c1c1c1c1c1}{1c1c1c1c1c1c1c1c1c1}
 \i & 4 	& \i	& \i & 1		&  5 		&\i &  2 		& \i & \i \\
 \i & 3 	& \i 	& \i &{\bf 0}& 4 		&\i &  1 		& \i & \i \\
 \i & 2 	& \i	& \i & 1		&  3 		&\i &{\bf 0}& \i & \i \\
 \i & 1 	& \i	& \i & 2	 	&  2 		&\i &{\bf 0}& \i & \i \\
 \i &{\bf 0}&\i	& \i & 3		&  1	 	&\i &  1 		& \i & \i \\
 \i & 1 	& \i	& \i & 4	 	&{\bf 0}&\i &  1 		& \i & \i \\
 \i & 2		& \i	& \i & 5	 	&  1 		&\i &{\bf 0}& \i & \i \\
 \i & 3 	& \i 	& \i & 6	 	&  2 		&\i &  1 		& \i & \i \\
 \i & 4 	& \i	& \i & 7	 	&  3 		&\i &  2 		& \i & \i 
\end{TAB}
}
	\hfil (c)
	{\scriptsize
\begin{TAB}(e,0.44cm,0.44cm){1c1c1c1c1c1c1c1c1c1c1}{1c1c1c1c1c1c1c1c1c1}
 5 &  4 	&  3	&  2 &  1		&   2 	& 3 &  2 		&  3 &  4 \\
 4 &  3 	&  2 	&  1 &{\bf 0}&  1 	& 2 &  1 		&  2 &  3 \\
 3 &  2 	&  3	&  2 & 1		&   2 	& 1 &{\bf 0}&  1 &  2 \\
 2 &  1 	&  2	&  3 & 2	 	&   2		& 1 &{\bf 0}&  1 &  2 \\
 1 &{\bf 0}& 1	&  2 & 2		&   1	 	& 2 &  1 		&  2 &  3 \\
 2 &  1 	&  2	&  2 & 1	 	&{\bf 0}& 1 &  1 		&  3 &  3 \\
 3 &  2		&  3	&  3 & 2	 	&  1 		& 1 &{\bf 0}&  1 &  2 \\
 4 &  3 	&  4 	&  4 & 3	 	&  2 		& 2 &  1 		&  2 &  3 \\
 5 &  4 	&  5	&  5 & 4	 	&  3 		& 3 &  2 		&  3 &  4 
\end{TAB}
}
	\caption{\label{fig_cityblockDTparallel}Parallel computation of city-block distances: (a) initialisation with zero distances at object-pixel positions; (b) calculation of vertical distances; (c) final result}
\end{figure}

An alternative approach is based on four scans, but with only one comparison per pixel access. It first processes all columns downward and upward: 
	\begin{align*}
		\begin{array}{clll}
		\mbox{downward scan in each column~} j: &
			\mbox{if} & d(i,j) > d(i-1,j) + 1 & d(i,j) := d(i-1,j)+1 \\
		\mbox{upward scan in each column~} j: &
			\mbox{if} & d(i,j) > d(i+1,j) + 1 & d(i,j) := d(i+1,j)+1 
		\end{array}
		\;.
	\end{align*}
Since the comparisons are independent of each other, all columns could be processed in parallel. Afterwards, all rows are processed in the same manner:
	\begin{align*}
		\begin{array}{clll}
		\mbox{forward scan in each row~} i: &
			\mbox{if} & d(i,j) > d(i,j-1) + 1 & d(i,j) := d(i,j-1)+1 \\
		\mbox{backward scan in each row~} i: &
			\mbox{if} & d(i,j) > d(i,j+1) + 1 & d(i,j) := d(i,j+1)+1 
		\end{array}
		\;.
	\end{align*}
\Listing{lst_cityblockDTpara} contains the corresponding source code and the example results are shown in \Figu{fig_cityblockDTparallel}. 
The final result is exactly the same as in Figure \ref{fig_cityblockDT}~(c).

\begin{figure}
	\hfil (a) \includegraphics[height=4.9cm]{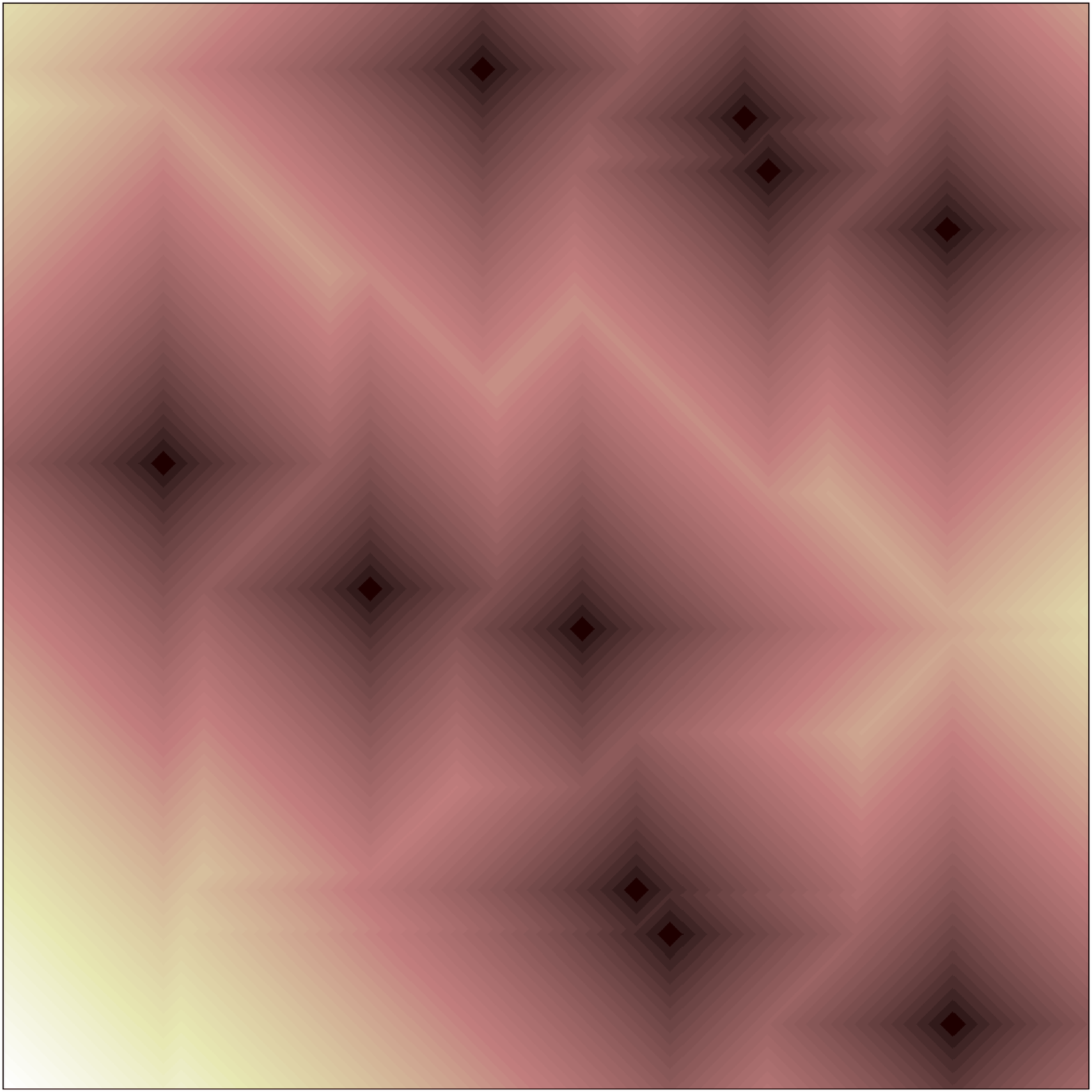}
	\hfil (b) \includegraphics[height=4.9cm]{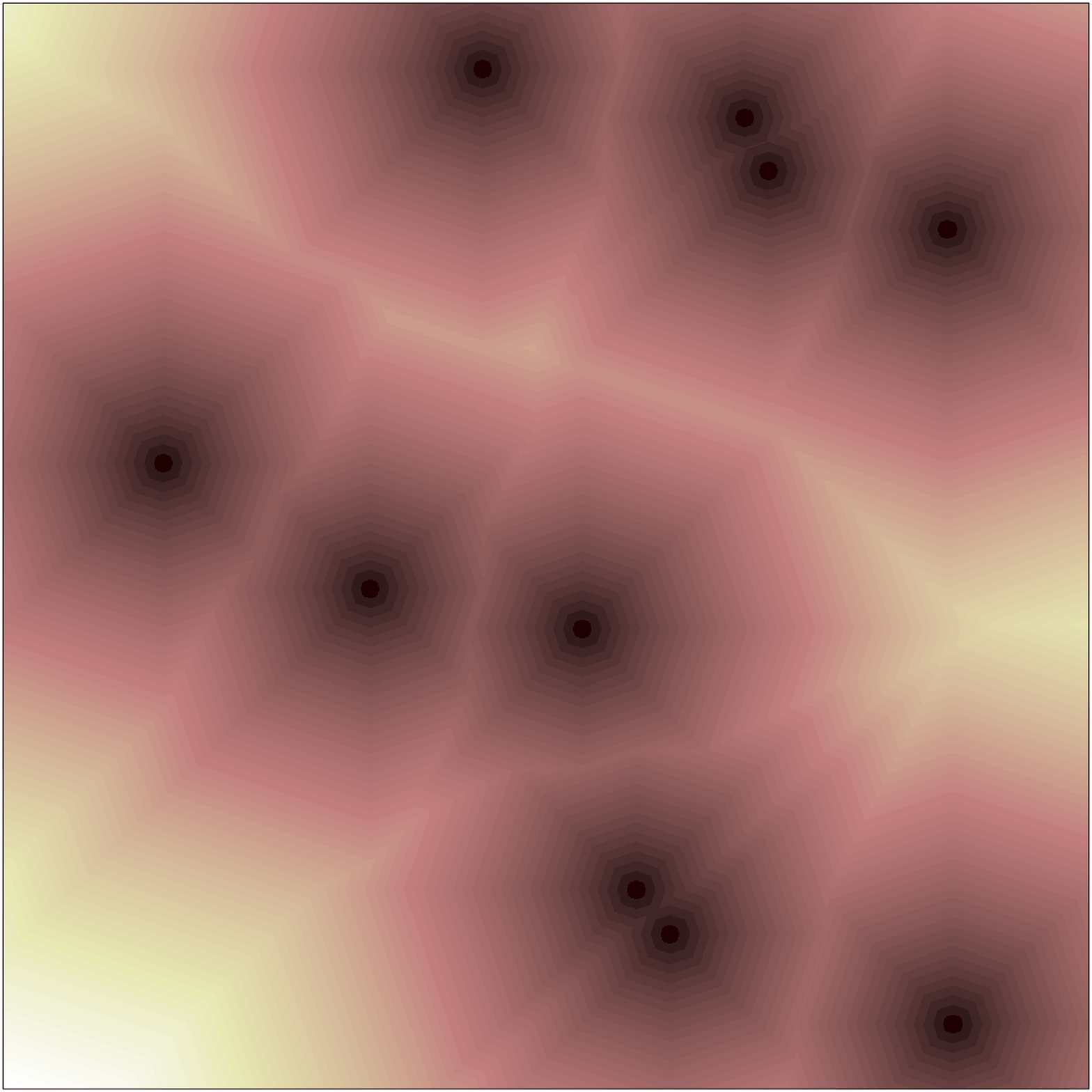}
	\hfil (c) \includegraphics[height=4.9cm]{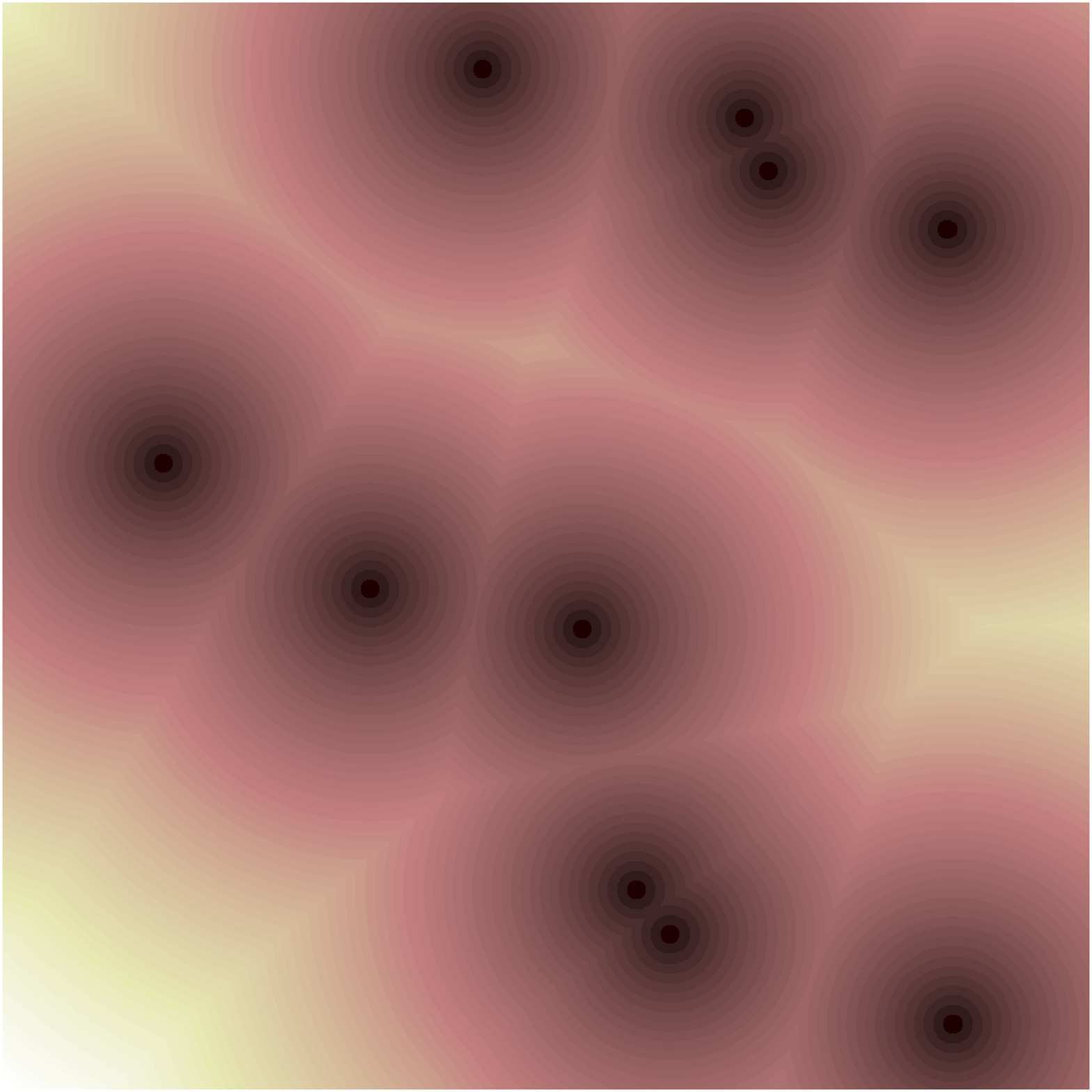}
	\caption{\label{fig_cityblockDTmat}Example of distance transforms based on: (a) city-block distance; (b) chamfer 4-3 distance; (c) propagation of relative positions. The brighter the pixels are, the longer is the distance to the closest object point.}
\end{figure}

\Figu{fig_cityblockDTmat}(a) depicts the distance transform result for a larger example with ten isolated object pixels. 
The brighter a pixel is in this picture, the higher is its distance to the closest object pixel. The typical diamond shape of the city-block distance propagation can be seen clearly (compare also Figure \ref{fig_cityblock}~b). 
 		%
\subsection{Chamfer distance transform}\label{subsec_chamferDT}
\begin{figure}
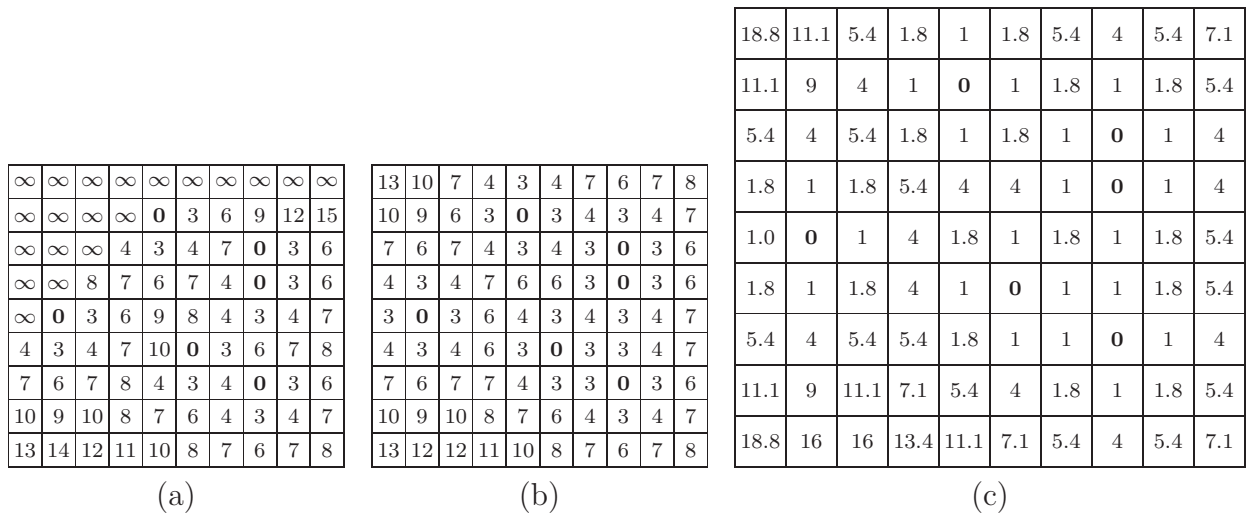

\hfil 
\parbox[t]{4.5cm}{
	{\scriptsize
\begin{TAB}(e,0.44cm,0.44cm){1c1c1c1c1c1c1c1c1c1c1}{1c1c1c1c1c1c1c1c1c1}
 \i & \i 	& \i	& \i & \i		& \i 		&\i & \i 		& \i & \i \\
 \i & \i 	& \i 	& \i &{\bf 0}&  3 	& 6 &  9 		& 12 & 15 \\
 \i & \i 	& \i	&  4 & 3		&   4 	& 7 &{\bf 0}&  3 &  6 \\
 \i & \i 	&  8	&  7 & 6	 	&   7		& 4 &{\bf 0}&  3 &  6 \\
 \i &{\bf 0}& 3	&  6 & 9		&   8	 	& 4 &  3 		&  4 &  7 \\
  4 & 3 	&  4	&  7 & 10	 	&{\bf 0}& 3 &  6 		&  7 &  8 \\
  7 & 6		&  7	&  8 & 4	 	&  3 		& 4 &{\bf 0}&  3 &  6 \\
 10 & 9 	& 10 	&  8 & 7	 	&  6 		& 4 &  3 		&  4 &  7 \\
 13 &14 	& 12	& 11 &10	 	&  8 		& 7 &  6 		&  7 &  8 
\end{TAB}
}~\centering{(a)}
}
\hfil
\parbox[t]{4.5cm}{
	{\scriptsize
\begin{TAB}(e,0.44cm,0.44cm){1c1c1c1c1c1c1c1c1c1c1}{1c1c1c1c1c1c1c1c1c1}
13 & 10 	&  7	&  4 &  3		&   4 	& 7 &  6 		&  7 &  8 \\
10 &  9 	&  6 	&  3 &{\bf 0}&  3 	& 4 &  3 		&  4 &  7 \\
 7 &  6 	&  7	&  4 &  3		&   4 	& 3 &{\bf 0}&  3 &  6 \\
 4 &  3 	&  4	&  7 &  6 	&   6		& 3 &{\bf 0}&  3 &  6 \\
 3 &{\bf 0}& 3	&  6 &  4		&   3	 	& 4 &  3 		&  4 &  7 \\
 4 &  3 	&  4	&  6 &  3 	&{\bf 0}& 3 &  3 		&  4 &  7 \\
 7 &  6		&  7	&  7 &  4 	&  3 		& 3 &{\bf 0}&  3 &  6 \\
10 &  9 	& 10 	&  8 &  7 	&  6 		& 4 &  3 		&  4 &  7 \\
13 & 12 	& 12	& 11 & 10 	&  8 		& 7 &  6 		&  7 &  8 
\end{TAB}
}~\centering{(b)}
}
\hfil
\parbox[t]{6.8cm}{
	{\scriptsize
\begin{TAB}(e,0.44cm,0.44cm){1c1c1c1c1c1c1c1c1c1c1}{1c1c1c1c1c1c1c1c1c1}
18.8 & 11.1	&  5.4	&  1.8 &  1		&   1.8	& 5.4 &  4 		& 5.4 & 7.1 \\
11.1 &  9 	&  4  	&  1	 &{\bf 0}&  1 	& 1.8 &  1 		& 1.8 & 5.4 \\
 5.4 &  4 	&  5.4	&  1.8 &  1		&   1.8	& 1	 	&{\bf 0}& 1		& 4	 \\
 1.8 &  1 	&  1.8	&  5.4 &  4 	&   4		& 1	 	&{\bf 0}& 1		& 4	 \\
 1.0 &{\bf 0}& 1  	&  4   &  1.8	&   1	 	& 1.8 &  1 		& 1.8 & 5.4 \\
 1.8 &  1 	&  1.8	&  4	 &  1 	&{\bf 0}& 1	 	&	 1 		& 1.8 & 5.4 \\
 5.4 &  4		&  5.4	&  5.4 &  1.8	&  1 		& 1	 	&{\bf 0}& 1		& 4		 \\
11.1 &  9 	& 11.1 	&  7.1 &  5.4	&  4 		& 1.8 &  1 		& 1.8 & 5.4 \\
18.8 & 16 	& 16		& 13.4 & 11.1	&  7.1	& 5.4 &  4 		& 5.4 & 7.1 
\end{TAB}
}~\centering{(c)}
}
	\caption{\label{fig_chamfer43DT}Computation of chamfer 4-3 distances: (a) propagation of distances from the top-left down to the bottom-right; (b) final result $d_{i,j}$ after backward propagation; (c) rounded values of $(d_{i,j}/3)^2$, integer numbers are exact values}
\end{figure}

In contrast to the city-block distance, the chamfer distance requires considerations in the 8-neighbourhood. Similar to what has been explained in Subsection \ref{subsec_cbDT}, also the chamfer distance can be computed in a sequential manner. Instead of two comparisons per scan and pixel, now four comparisons are needed. The scan from top-left to bottom right makes following decisions:
	\begin{align*}
		\mbox{for all} (i,j)\left\{
		\begin{array}{lll}
			\mbox{if} & d(i,j) > d(i,j-1) + 3 & d(i,j) := d(i,j-1)+3 \\
			\mbox{if} & d(i,j) > d(i-1,j) + 3 & d(i,j) := d(i-1,j)+3 \\
			\mbox{if} & d(i,j) > d(i-1,j-1) + 4 & d(i,j) := d(i-1,j-1)+4 \\
			\mbox{if} & d(i,j) > d(i-1,j+1) + 4 & d(i,j) := d(i-1,j+1)+4 \\
		\end{array}
		\right.
		\;.
	\end{align*}
The backward scan is accordingly 
	\begin{align*}
		\mbox{for all} (i,j)\left\{
		\begin{array}{lll}
			\mbox{if} & d(i,j) > d(i,j+1) + 3 & d(i,j) := d(i,j+1)+3 \\
			\mbox{if} & d(i,j) > d(i+1,j) + 3 & d(i,j) := d(i+1,j)+3 \\
			\mbox{if} & d(i,j) > d(i+1,j-1) + 4 & d(i,j) := d(i+1,j-1)+4 \\
			\mbox{if} & d(i,j) > d(i+1,j+1) + 4 & d(i,j) := d(i+1,j+1)+4 \\
		\end{array}
		\right.
		\;.
	\end{align*}
\Figu{fig_chamfer43DT}(a)+(b) depict the intermediate and the final result.
The third matrix in Figure \ref{fig_chamfer43DT}(c) also contains the final result including a normalization with three and subsequent squaring. 
This modification scales the values back to a horizontal distance equal to one between two pixels and allows the comparison with exact Euclidean distances in Figure \ref{fig_EEDT}(c).

The shape of the distance propagation using chamfer distances has changed to an octagon, see \Figu{fig_cityblockDTmat}(b) .
 		%
\subsection{Approximate Euclidean Distance Transform}
\begin{figure}
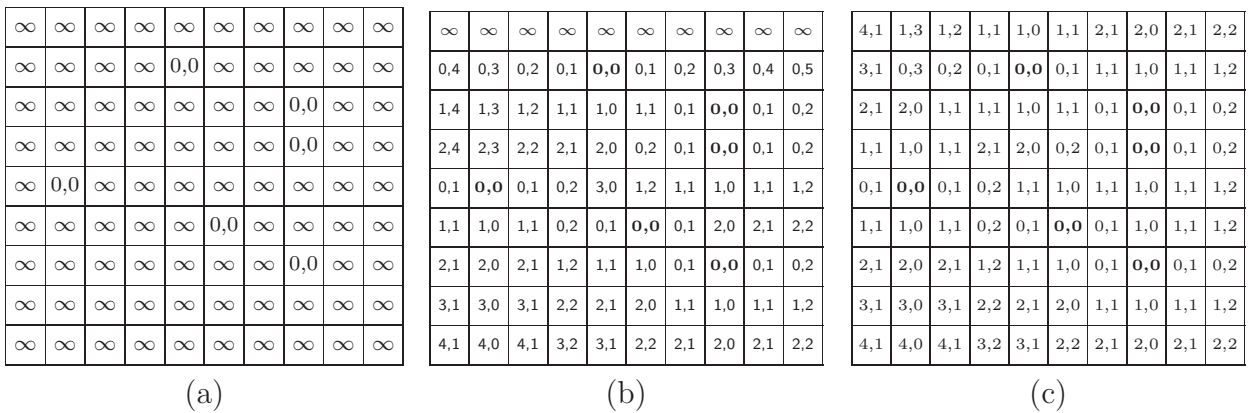

\hfil 
\parbox[t]{5.3cm}{
	{\scriptsize
\begin{TAB}(e,0.3cm,0.3cm){1c1c1c1c1c1c1c1c1c1c1}{1c1c1c1c1c1c1c1c1c1}
 \i & \i 	& \i	& \i & \i & \i &\i & \i & \i & \i \\
 \i & \i 	& \i 	& \i & 0,0& \i &\i & \i & \i & \i \\
 \i & \i 	& \i	& \i & \i & \i &\i &0,0 & \i & \i \\
 \i & \i 	& \i	& \i & \i & \i &\i &0,0 & \i & \i \\
 \i & 0,0	& \i	& \i & \i	& \i &\i & \i & \i & \i \\
 \i & \i 	& \i	& \i & \i &0,0 &\i & \i & \i & \i \\
 \i & \i	& \i	& \i & \i & \i &\i &0,0 & \i & \i \\
 \i & \i 	& \i 	& \i & \i & \i &\i & \i & \i & \i \\
 \i & \i 	& \i	& \i & \i & \i &\i & \i & \i & \i 
\end{TAB}
}~\centering{(a)}
}
\hfil 
\parbox[t]{5.3cm}{
	{\sf\tiny
\begin{TAB}(e,0.4cm,0.4cm){1c1c1c1c1c1c1c1c1c1c1}{1c1c1c1c1c1c1c1c1c1}
 \i 	& \i 			& \i	& \i & \i 		& \i 			&\i 	& \i 			& \i & \i \\
 0,4 	& 0,3			&0,2 	&0,1 &{\bf 0,0}&0,1 		&0,2	&0,3 			&0,4 &0,5 \\
 1,4 	& 1,3 		&1,2	&1,1 &1,0 		&1,1 			&0,1	&{\bf 0,0}&0,1 &0,2 \\
 2,4	& 2,3			&2,2	&2,1 &2,0 		&0,2 			&0,1	&{\bf 0,0}&0,1 &0,2 \\
 0,1	&{\bf 0,0}&0,1	&0,2 &3,0			&1,2 			&1,1	&1,0			&1,1 &1,2 \\
 1,1 & 1,0 			& 1,1	&0,2 &0,1 		&{\bf 0,0}&0,1	&2,0 			&2,1 & 2,2 \\
 2,1 & 2,0 			& 2,1	&1,2 &1,1 		&1,0 			&0,1	&{\bf 0,0}&0,1 & 0,2 \\
 3,1 & 3,0 			& 3,1	&2,2 &2,1 		&2,0 			&1,1	&1,0 			&1,1 & 1,2 \\
 4,1 & 4,0 			& 4,1	&3,2 &3,1 		&2,2 			&2,1	&2,0 			&2,1 & 2,2 \\
\end{TAB}
}~\centering{(b)}
}
\hfil 
\parbox[t]{5.3cm}{
	{\tiny
\begin{TAB}(e,0.4cm,0.4cm){1c1c1c1c1c1c1c1c1c1c1}{1c1c1c1c1c1c1c1c1c1}
 4,1	& 1,3 		& 1,2	&1,1 &1,0 		&1,1 			&2,1 &2,0 		&2,1	&2,2 \\
 3,1	& 0,3 		& 0,2	&0,1 &{\bf 0,0}&0,1 		&1,1 &1,0 		&1,1	&1,2 \\
 2,1 	& 2,0 		& 1,1	&1,1 &1,0 		&1,1 			&0,1 &{\bf 0,0}&0,1 &0,2 \\
 1,1 	& 1,0 		& 1,1	&2,1 &2,0 		&0,2 			&0,1 &{\bf 0,0}&0,1 &0,2 \\
 0,1 	&{\bf 0,0}& 0,1	&0,2 &1,1			&1,0 			&1,1 &1,0 		&1,1	&1,2 \\
 1,1	& 1,0 		& 1,1	&0,2 &0,1 		&{\bf 0,0}&0,1 &1,0 		&1,1	&1,2 \\
 2,1	& 2,0 		& 2,1	&1,2 &1,1 		&1,0 			&0,1 &{\bf 0,0}&0,1 &0,2 \\
 3,1	& 3,0 		& 3,1	&2,2 &2,1 		&2,0 			&1,1 &1,0 		&1,1	&1,2 \\
 4,1	& 4,0 		& 4,1	&3,2 &3,1 		&2,2 			&2,1 &2,0 		&2,1	&2,2 \\
\end{TAB}
}~\centering{(c)}
}
	\caption{\label{fig_DanielssonDT}Approximate Euclidean distance transform: (a) initialisation of object pixel positions; (b) propagated relative positions $(\Delta_y,\Delta_x)$ after first scan from top to bottom; (c) final result $(\Delta_y,\Delta_x)$ after propagation from bottom to top; }
\end{figure}

Danielsson had the idea to not propagate the distances but the relative vertical and horizontal positions \cite{Dan80}. Object pixels always have a relative position of $[\Delta_y, \Delta_x]=[0,0]$, see \Figu{fig_DanielssonDT}(a).
Each horizontal step increments the horizontal relative position $\Delta_x$ and each vertical step increments the vertical relative position $\Delta_y$. A diagonal step increments both. The Euclidean distance can be computed with $d = (\Delta_y^2+\Delta_x^2)^{0.5}$. 

The source code in \Listing{lst_DanielssonDT}, \Listing{lst_DanielssonDT2}, and \Listing{lst_DanielssonDT3} differs somewhat from Danielsson's proposal and follows the modifications of \cite{Grevera2007}.

In addition to the distance matrix {\tt distMat} that has to be generated, two other matrices {\tt distMatY} and {\tt distMatX} have to be maintained. They store the propagated relative positions. 
Based on these relative positions, the squared Euclidean distances have to be determined. Instead of computing these distances again and again, a function {\tt getDist()} is used. It calculates the different distances only once, stores these values in a matrix {\tt preCalc}, and also utilizes the symmetric property of the metric.

The source code in Listings \ref{lst_DanielssonDT2} and \ref{lst_DanielssonDT3} is somewhat lengthy, since several decisions have to be made. 
The processing comprises two scans: the first is starting in the second row going down to the bottom row and the second is operating in opposite direction.
In each scan, the rows are processed forward and backward.
The lines 3-14 in Listing \ref{lst_DanielssonDT2}, for example, perform a vertical propagation of relative positions.
The values from the top neighbour position are taken ({\tt yr = distMatY(i-1,j); xr = distMatX(i-1,j);}). The vertical component is incremented ({\tt yr+1}) and the corresponding squared distance is determined using the function {\tt getDist()}.
If the current distance is longer than this new one ({\tt if distMat(i,j) > dist}), it is replaced and also the relative positions are updated ({\tt distMat(i,j) = dist; distMatY(i,j) = yr+1; distMatY(i,j) = xr}).
However, function {\tt getDist()} may not be accessed when the relative positions are larger than the dimensions of matrix {\tt preCalc} ({\tt if yr < maxDist}).

This procedure is repeated for the two other positions in the causal neighbourhood {\tt (i,j-1)} and {\tt (i-1,j-1)}. The subsequent backward scan for this row checks the positions {\tt (i,j+1)} and {\tt (i-1,j+1)}.
\Figu{fig_DanielssonDT}(b) shows the result of this first scan. 
Note that the sorted pairs $a,b$ and $b,a$ correspond to the same distance $d^2=a^2+b^2$. However, the derived order of $a$ and $b$ indicates to which object pixel each background pixel has been assigned.

The second scan propagates the minimum distances from the bottom row upwards (Listing \ref{lst_DanielssonDT3}) in a very similar manner and produces the final result (\Figu{fig_DanielssonDT}~c).
For the given example, this corresponds to the exact Euclidean distances as shown in Figure \ref{fig_EEDT}. 

\begin{figure}
	 (a)\includegraphics[scale=0.6]{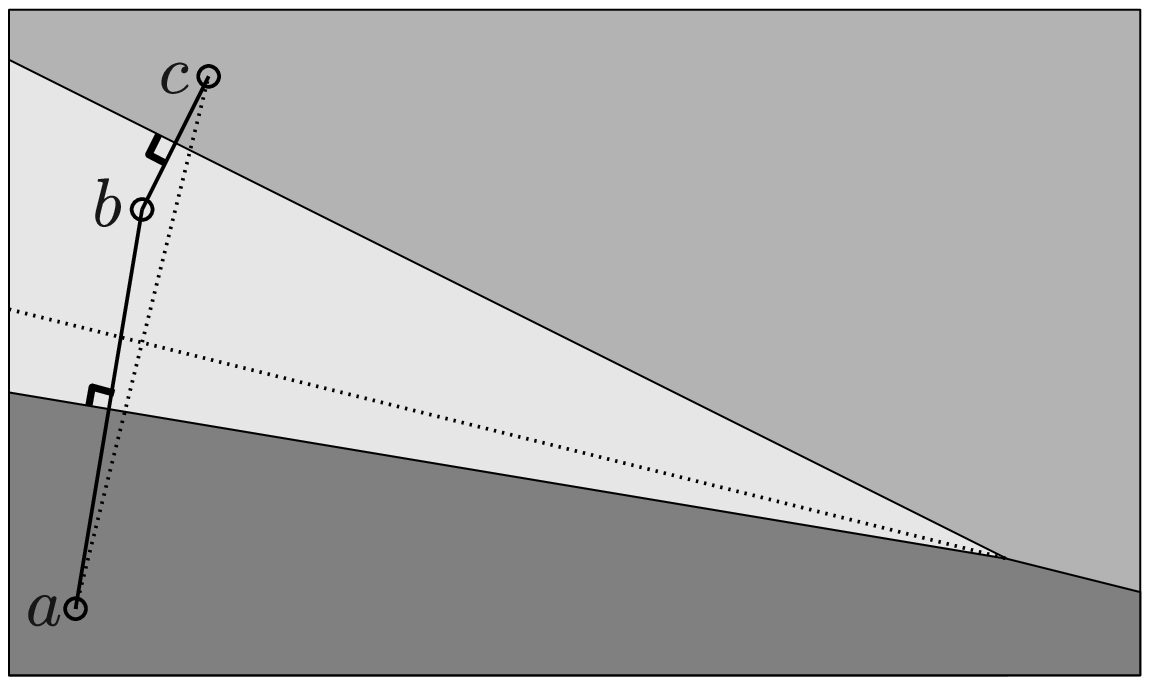}
	\hfil (b)\includegraphics[scale=0.75]{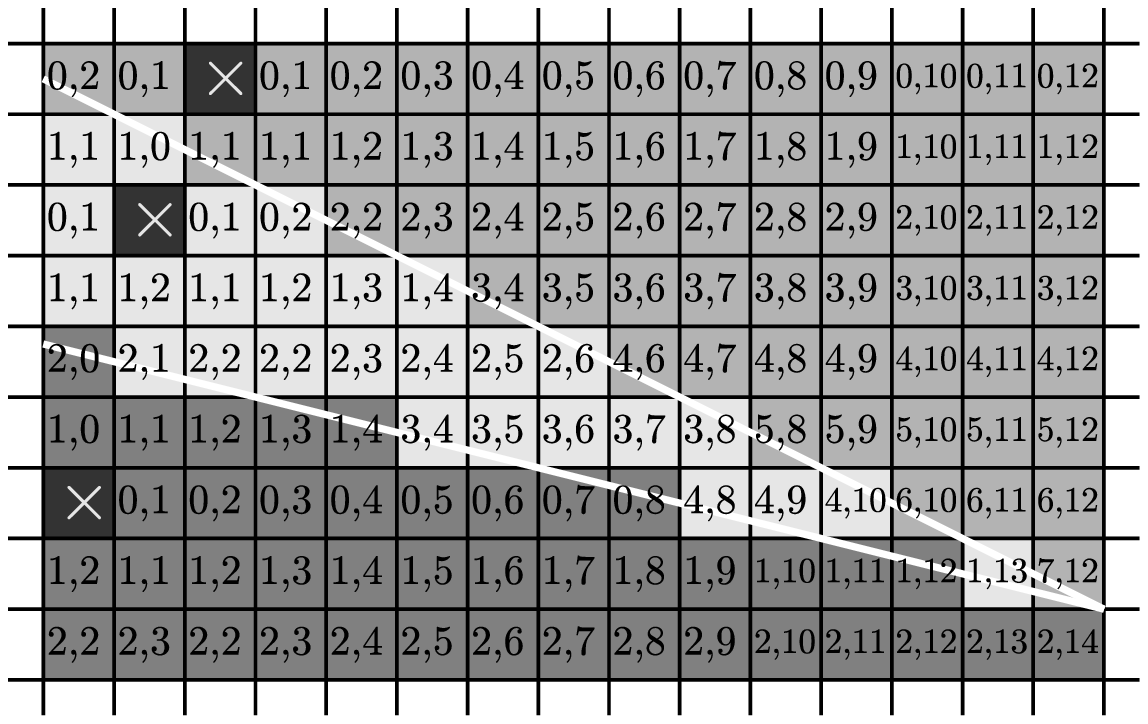}
	\caption{\label{fig_propError}Assignment of points: (a) continuous space separated in three regions which are closest to either point $a$, $b$, or $c$; (b) example result of Danielssons approach. Object pixels are marked with $\times$. The bright isolated pixel has been assigned to the wrong object pixel with a distance of $d^2=1^2+13^2 =170$. However, its correct distance is $d^2=5^2+12^2 =169$ with relation to the object pixel in the middle.}
\end{figure}

Under special circumstances, it is impossible to derive the correct distance. This is because the algorithm depends on the propagation of distances, while the assignment of the closest object pixel can lead to disconnected background pixels. This has already been mentioned and explained by Danielsson himself \cite{Dan80} and different numeric examples have been given in \cite{Bai04} and \cite{Fab08}.
One example can be seen in \Figu{fig_propError}.
The continuous space is separated in regions by drawing the	perpendicular bisector between two points. There is a clear assignment of each position to one of the three point, Figure \ref{fig_propError}(a). In discrete space, 
the three object pixels (marked with a cross) also split the set of background pixels in three region, Figure \ref{fig_propError}(b). The white lines show the separation in continuous space. The discretization causes a disruption of the middle region. 
The approach of Danielsson is not able to take account of this disruption and assigns the isolated pixel to the bottom region with a relative position of $1,13$. There is no chance to propagate the positions from `4,10' to the correct relative distance of `5,12' via `5,11' or via `4,11' because the corresponding positions are closer to the other object pixels.

A typical visual appearance of this distance transform result is shown in \Figu{fig_cityblockDTmat}(c). It looks in principle like a result produced by an exact Euclidean distance transform.

		%
\section{Exact Euclidean Distance Transform}
\label{sec_EEDT}
    %
Some applications depend on the exact calculation of the distances between each background pixel and the nearest object pixel. Thus, the following explanations refer to the case shown in Figure \ref{fig_examples}~(c).
In principle, one could first identify the inner contours of all objects. The inner contour includes all object pixels that have at least one background pixel in their neighbourhood.
In a second step, one could compute the Euclidean distances from each background pixel to all contour pixels and find the minimum. This minimum distance is then stored at the position of the examined background pixel.

In worst case, each object pixel could be a contour pixel\footnote{For example, each object could be represented by a single isolated pixel.}. Let $n_O$ denote the number of contour pixels. The total number of pixels is given by the image size $L\times M$. The number of background pixels is then $n_B= L\cdot M - n_O$. The number $n_c$ of distance calculations would be 
	\begin{align}
	  n_c = n_B \cdot n_O = (L\cdot M - n_O) \cdot n_O
	\end{align}
The maximum of required calculations $\max(n_c) = (L\cdot M)^2/4$ is reached for $n_B = n_O = (L\cdot M)/2$.
Consequently, this naive approach has a quadratic complexity with respect to the image size: $O((LM)^2)$.

For a better understanding of the following text, the most important variables are summarized in \Table{tab_notation}.
\begin{table}
	\caption{\label{tab_notation}Notation}
\hfil
\framebox{
\vbox{
\begin{tabbing}
 ww\= WW\= qwww\kill
$d$ 	\> \dots	\> distance value \\
$D$ 	\> \dots	\> squared distance $D=d^2$ \\
$\Delta$ 	\> \dots	\> difference between elements of coordinates \\
$i$ 	\> \dots	\> row index \\
$i_0$ \> \dots	\> index of current or particular row \\
$i_n$ \> \dots	\> vertical coordinate of an object pixels ${\mathbf p}_n$ \\
$j$ 	\> \dots	\> column index\\
$j_0$ \> \dots	\> index of current or particular column \\
$j_n$ \> \dots	\> horizontal coordinate of an object pixels ${\mathbf p}_n$ \\
$j_s$ \> \dots	\> horizontal location of an intersection, $j_s\in\setR$\\ 
$k$ 	\> \dots	\> column index related to object pixels\\
$L$ 	\> \dots	\> number of columns \\
$M$ 	\> \dots	\> number of rows \\
$n$ 	\> \dots	\> index for object pixels \\
${\mathbf p}_n$ 	\> \dots	\> object pixel
\end{tabbing}
}}
\end{table}
	%
		%
\subsection{Basic approach}\label{subsec_basic}
\begin{figure}
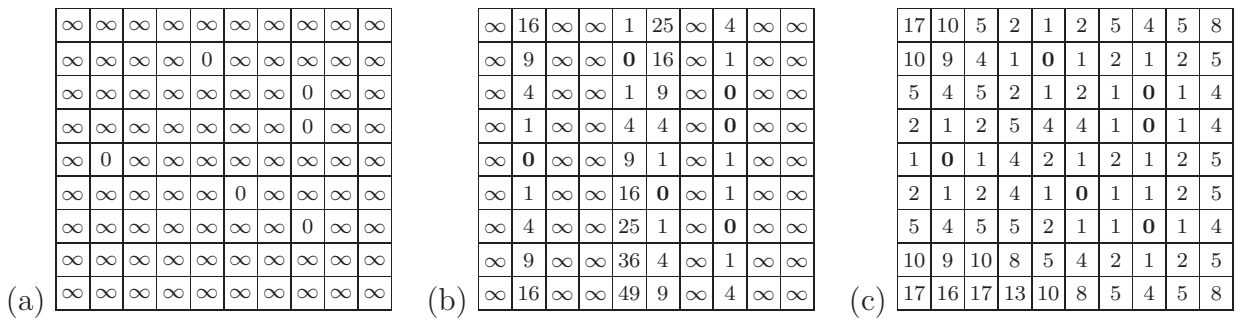

	\hfil (a)
	{\scriptsize
\begin{TAB}(e,0.44cm,0.44cm){1c1c1c1c1c1c1c1c1c1c1}{1c1c1c1c1c1c1c1c1c1}
 \i & \i 	& \i	& \i & \i & \i &\i & \i & \i & \i \\
 \i & \i 	& \i 	& \i & 0 	& \i &\i & \i & \i & \i \\
 \i & \i 	& \i	& \i & \i & \i &\i &  0 & \i & \i \\
 \i & \i 	& \i	& \i & \i & \i &\i &  0 & \i & \i \\
 \i & 0 	& \i	& \i & \i	& \i &\i & \i & \i & \i \\
 \i & \i 	& \i	& \i & \i &  0 &\i & \i & \i & \i \\
 \i & \i	& \i	& \i & \i & \i &\i &  0 & \i & \i \\
 \i & \i 	& \i 	& \i & \i & \i &\i & \i & \i & \i \\
 \i & \i 	& \i	& \i & \i & \i &\i & \i & \i & \i 
\end{TAB}
}
	\hfil (b)
	{\scriptsize
\begin{TAB}(e,0.44cm,0.44cm){1c1c1c1c1c1c1c1c1c1c1}{1c1c1c1c1c1c1c1c1c1}
 \i & 16 	& \i	& \i & 1		& 25 		&\i &  4 		& \i & \i \\
 \i & 9 	& \i 	& \i &{\bf 0}& 16 	&\i &  1 		& \i & \i \\
 \i & 4 	& \i	& \i & 1		&  9 		&\i &{\bf 0}& \i & \i \\
 \i & 1 	& \i	& \i & 4	 	&  4 		&\i &{\bf 0}& \i & \i \\
 \i &{\bf 0}&\i	& \i & 9		&  1	 	&\i &  1 		& \i & \i \\
 \i & 1 	& \i	& \i & 16	 	&{\bf 0}&\i &  1 		& \i & \i \\
 \i & 4		& \i	& \i & 25	 	&  1 		&\i &{\bf 0}& \i & \i \\
 \i & 9 	& \i 	& \i & 36	 	&  4 		&\i &  1 		& \i & \i \\
 \i & 16 	& \i	& \i & 49	 	&  9 		&\i &  4 		& \i & \i 
\end{TAB}
}
	\hfil (c)
	{\scriptsize
\begin{TAB}(e,0.44cm,0.44cm){1c1c1c1c1c1c1c1c1c1c1}{1c1c1c1c1c1c1c1c1c1}
 17 & 10 	&  5	& 2 & 1		& 2 		&5 &  4 		& 5 & 8 \\
 10 & 9 	&  4 	& 1 &{\bf 0}& 1 	&2 &  1 		& 2 & 5 \\
  5 & 4 	&  5	& 2 & 1		&  2 		&1 &{\bf 0} & 1 & 4 \\
  2 & 1 	&  2	& 5 & 4	 	&  4 		&1 &{\bf 0} & 1 & 4 \\
  1 &{\bf 0}&1	& 4 & 2		&  1	 	&2 &  1 		& 2 & 5 \\
  2 & 1 	&  2	& 4 & 1	 	&{\bf 0}&1 &  1 		& 2 & 5 \\
  5 & 4		&  5	& 5 & 2	 	&  1 		&1 &{\bf 0} & 1 & 4 \\
 10 & 9 	& 10 	& 8 & 5	 	&  4 		&2 &  1 		& 2 & 5 \\
 17 & 16 	& 17	&13 & 10 	&  8 		&5 &  4 		& 5 & 8 
\end{TAB}
}
	\caption{\label{fig_EEDT}Computation of squared Euclidean distances: (a) initialisation with zero distances at object-pixel positions; (b) calculation of vertical distances; (c) final result}
\end{figure}

Many researchers have thought of algorithms that can compute the exact Euclidean distance transform (EEDT) in a faster manner.
Some of these successful methods are based on the separate processing of all dimensions \cite{Mei00,Mau03,Fel12}. 
Let us take an example with six object pixels. The distance matrix in \Figu{fig_EEDT}(a) shows the positions of object pixels with a value of zero. 
The following derivations use squared Euclidean distances 
	\begin{align}\label{eq_squaredEuclid}
	  D= d^2 = (p_x-q_x)^2 + (p_y-q_y)^2 
		\;,
	\end{align}
with $\mathbf{p}=(p_x, p_y)$ being the position of a background pixel and $\mathbf{q}=(q_x, q_y)$ being the position of an object pixel. The square root can be drawn (if necessary) after all squared distances have been determined.

The distances for the background pixels are not known yet and they are initialised with a sufficiently large value higher than $(M-1)^2 + (L-1)^2$, which represents the maximum possible distance for a $M\times L$ matrix. In Figure \ref{fig_EEDT}, this value is indicated by `$\infty$'.

From (\ref{eq_squaredEuclid}) it can be seen that the distance consists of two additive terms, which can be computed independently. For example, one could first calculate all vertical components $D_{\rm v}=(p_y-q_y)^2$. The result is depicted in \Figu{fig_EEDT}(b).

In a second step, the horizontal components $D_{\rm h}=(p_x-q_x)^2$ have to be determined row by row.
At all positions in each row, it has to be checked which combination $D=D_{\rm v}+D_{\rm h}$ of vertical and horizontal components results in the shortest distance. Starting in the top-left corner, there is no distance assigned yet. The row is scanned to the right for the next available distance component. A value of $D_{\rm v}=16$ can be found at a horizontal distance of $d_{\rm h}=1$. The combination of both components is $D_2 = 1^2 + 16=17$. The index $k$ in $D_k$ indicates the column of the used vertical distance. This is the first candidate. The second can be computed at a horizontal distance of 4: $D_5 = 4^2 + 1=17$. Two more candidates are available: $D_6 = 5^2 + 25=50$ and $D_8 = 7^2 + 4=53$.
The best candidate is obviously
	\begin{align*}
	  D^* = \min\limits_{i}(D_i) = 17
		\;.
	\end{align*}
This value is written to the current pixel position.
The candidates for the second pixel in the first row are: 
	\begin{align*}
	  D_2 &= 0^2 + 16=16 \quad\mbox{(pure vertical distance)}\\
	  D_5 &= 3^2 + ~1=10 \\
	  D_6 &= 4^2 + 25=41 \\
	  D_8 &= 6^2 + ~4=40
		\;.
	\end{align*}
The already existing value of 16 has to be replaced by $D^*=10$, because this is the shortest distance to any object pixel. This procedure continues until all distances are fixed for this row. Then all other rows are processed in the same manner.

\Listing{lst_eedtSimple} contains a possible algorithm for this procedure. This source code allows an estimation of the corresponding complexity. Lines 15 to 34 compute the vertical components. Interestingly, there is no multiplication involved despite the fact that $(p_y-q_y)\cdot (p_y-q_y)$ has to be computed for each background pixel. The outer loop {\tt for j=1:L} steps through all columns. There are two inner loops:  {\tt for i=2:M} and {\tt for i=M-1:-1:1}. The first one scans the current column downwards starting with the second row and the second loop scans the same column upwards.
Both try to propagate distances in such a manner that larger distance entries are overwritten by possibly smaller ones.

With respect to the example in Figure \ref{fig_EEDT}(a), the downward scan firstly enfolds its impact in the second column (counted from the left) at the position right under the object pixel with distance equal to zero. At this position it finds a distance value that is larger ($=\infty$) than the distance above plus the actual distance step, which had been initialized with {\tt distStep=1}. The current value is set to $0+1=1$ and the distance step is increased by two. What's the deal with that increment?
	
	Given two consecutive numbers $x$ and $x+1$, the difference between their squared values is $(x+1)^2 - x^2 = 2x+1$. Starting with $x=0$, the first difference is equal to one ({\tt distStep=1}). Then the difference between the squared values grows by 2 ({\tt distStep = distStep + 2;}) with each increment of $x$. Using this step as an additive term in {\tt distMat(i,j) = distMat(i-1,j) + distStep}, the algorithm computes the required squared distance values without any multiplication.
	
	As soon as the position of an object pixel is reached, the current distance value is not larger than the propagated one; the distance value remains unchanged and {\tt distStep} is reset to one.
	The result of the downwards propagation is shown in \Figu{fig_EEDT2}.
\begin{figure}
\hfil
	{\scriptsize
\begin{TAB}(e,0.44cm,0.44cm){1c1c1c1c1c1c1c1c1c1c1}{1c1c1c1c1c1c1c1c1c1}
 \i & \i 	& \i	& \i & \i		& \i 		&\i &  \i		& \i & \i \\
 \i & \i 	& \i 	& \i &{\bf 0}& \i 	&\i &  \i		& \i & \i \\
 \i & \i 	& \i	& \i & 1		&  \i		&\i &{\bf 0}& \i & \i \\
 \i & \i 	& \i	& \i & 4	 	&  \i		&\i &{\bf 0}& \i & \i \\
 \i &{\bf 0}&\i	& \i & 9		&  \i 	&\i &  1 		& \i & \i \\
 \i & 1 	& \i	& \i & 16	 	&{\bf 0}&\i &  4 		& \i & \i \\
 \i & 4		& \i	& \i & 25	 	&  1 		&\i &{\bf 0}& \i & \i \\
 \i & 9 	& \i 	& \i & 36	 	&  4 		&\i &  1 		& \i & \i \\
 \i & 16 	& \i	& \i & 49	 	&  9 		&\i &  4 		& \i & \i 
\end{TAB}
}
	\caption{\label{fig_EEDT2}Result after downward propagation of squared Euclidean distances, see example from Figure \ref{fig_EEDT}(a)}
\end{figure}
The upward scan within a column does not change anything before the first object pixel is passed. After this position the distances are updated if they are larger than the propagated values, see Figure \ref{fig_EEDT}(b).
These scans are rather fast and there is only opportunity for improvements in the upward scan, if there are many columns without any object pixel (entire column could be skipped in the upward scan) or the object pixels concentrate in the upper rows of the image (lower part of the column could be skipped). In maximum there are $L\cdot M$ distances to be computed. So, the vertical propagation has linear complexity with respect to the image size.

The processing of rows needs more efforts and defines the complexity of the entire approach, Listing \ref{lst_eedtSimple} lines 37-52.
The outer loop {\tt for i=1:M} takes care of all matrix rows. The corresponding row of vertical distances is copied to a new vector {\tt distMatV} because the values in matrix {\tt distMat} can be overwritten. In lines 40 and 43, there are two nested inner loops, both scanning all positions within the current row. Index $j$ of the first inner loop defines the position for which the minimum distance has to be found. The second inner loop scans for all possible candidates of vertical distances {\tt distMatV(k)}.

\begin{figure}
	\hfil \includegraphics[scale=0.6]{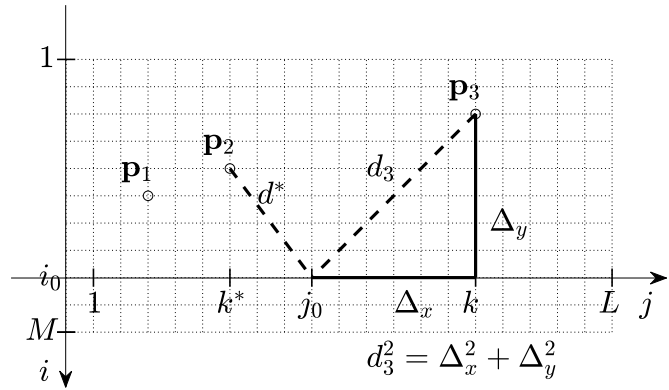}
	\caption{\label{fig_diagram}Looking for closest object point $p_k$ for current background pixel at position $(i_0, j_0)$. The current minimum distance is $D^*$ related to point ${\mathbf p}_2$ from column $k^*$. The distance measure $d_3$ cannot be shorter because its horizontal component $\Delta_x$ is already longer than the current minimum: $\Delta_x > d^*$.}
\end{figure}

\Figu{fig_diagram} shows a situation where a background pixel at position $(i_0, j_0)$ has to be assigned the distance to the closest object pixel $\mathbf{p_k}$.
Scanning the line $i_0$ from $k=1$ towards $k=L$, the current point has already been compared to the points $p_1$ and $p_2$ resulting to a minimum distance of $d^* = \sqrt{4^2 + 3^2} = 5$ in this example. The next vertical distance $D_{\rm v} < \infty$ can be found in the column of point $\mathbf{p_3}$.

In the worst case, all positions $j_0$ have to be compared with all positions $k$ with $1\le j_0,k\le L$, i.e.\ $L\cdot (L-1)$ distances have to be computed. Taking all rows into account, we have $M\cdot L\cdot (L-1)$ distances to calculate leading to a $O(ML^2)$ complexity, which is at least somewhat lower than for the naive approach mentioned in the beginning of Section \ref{sec_EEDT}.
		%
\subsection{Faster approaches}
The most important question is, whether it is possible to skip some of the distance calculations.
While the determination of vertical distances is already quite fast and has only linear $O(M)$ complexity, the calculation of distances along the rows has some room for optimizations.
 With respect to the basic approach discussed above, there are some tricks to speed up the processing.
		%
\subsubsection{Change order of column and row processing}
Depending on the aspect ratio of the given image, the complexity $O(LM^2)$ is lower than $O(ML^2)$ if $M<L$. If the complexity of the row-wise processing will not reduced by another mechanism, the order of processing columns and rows should be exchanged in this case. 
		%
\subsubsection{Avoid unnecessary computations}
The computations in lines 43-49 of Listing \ref{lst_eedtSimple} should only be executed if there is a chance to find a new minimum distance. Columns without any object point contain infinite ({\tt =maxDist}) vertical distances. So, checking `{\tt if distMatV(k) < maxDist}', it can be ensured that variable $k$ is not pointing at such column and unnecessary computations can be avoided. However, if most of the columns are occupied by at least one object pixel, this has little effect.

A second simple trick that can be combined with the first one concerns the horizontal component of the distances to be computed. 

As soon as a position $k= j_0 + \Delta_x$ with $\Delta_x > d^*$ is reached, no shorter path than $d^*$ can be found and thus the scan can be terminated. 
This situation is shown in Figure \ref{fig_diagram}. The current minimum distance $d^*$ refers to point $\mathbf{p_2}$ and the next available point $\mathbf{p_3}$ has a horizontal distance that already is larger than $d^*$ and the processing of loop {\tt for k=1:L} can be stopped.
This leads to large saving effects when $j_0$ is small. However, the effect decreases the more the current position $j_0$ moves towards $L$.
If the scanning process is split into a forward scan $j_0<k\le L$, which can be processed as described and a backward scan from $k=j_0-1$ to $k=1$, then it is possible to stop both scanning processes as soon as $|k-j_0|\ge d^*$ is reached.
These modifications are implemented in \Listing{lst_eedtImproved}. Remember that all distances are squared ones.
		%
\subsection{Fast exact Euclidean distance transforms}
The previous Subsections explained how the exact Euclidean distance transform can be realised and they proposed to skip unnecessary computations. However, the discussed methods for doing so are quite simple and have impact only under special conditions. 
As mentioned in Subsection \ref{subsec_basic}, only the vertical distance component $D_{\rm v}$ can be determined in linear time ($O(M)$) while the horizontal distance component $D_{\rm h}$ still needs $O(L^2)$ leading to a total complexity of $O(ML^2)$.
However, based on deeper considerations, it is possible to reduce the complexity to $O(ML)$ as different scientists have proven. With other words: also the determination of horizontal distances can be performed in linear time.

The explanations in Subsection \ref{subsec_basic} regarding the combination of the vertical and the horizontal component to find the final distance in (\ref{eq_squaredEuclid}) can be viewed from another perspective.
Each object point $\mathbf{p}_n$ located at $(i_n, j_n)$ contributes to row $i_0$ with a vertical component $D_{n,{\rm v}}(i_0) = (i_n - i_0)^2$. According to Figure \ref{fig_diagram}, its horizontal component is $D_{n,{\rm h}}(j_0) = (j_n - j_0)^2$. 
For each point $\mathbf{p}_n$ and a given row $i_0$, all distances $D_n(i_0,j_0) = D_{n,{\rm v}}(i_0) + D_{n,{\rm h}}(j_0)$ for all horizontal positions $j_0$, with $1\le j_0\le L$ could be computed.
With respect to the example shown in Figure \ref{fig_EEDT}~(b), we could take the first row containing four vertical components $\{D_{n,{\rm v}}(i_0=1)\}=\{16, 1, 25, 4\}$ and could complement them separately with all possible horizontal components $D_{n,{\rm h}}(j_0)$. The result is shown in \Figu{fig_propagation}.
\begin{figure}
\hfil
	{\scriptsize
\begin{TAB}(e,0.44cm,0.44cm){c1c1c1c1c1c1c1c1c1c1c1}{1c1c1c1c1c1c1c1c}
 $j,k$	&  1 	& 2 			& 3 	&  4 	& 5			& 6 			&	 7	&  8		&  9 	& 10 \\
$D_{k,{\rm v}}$
				& \i 	&{\bf 16}	& \i 	& \i 	&{\bf 1}&{\bf 25}	& \i 	&{\bf 4}& \i	& \i \\
$D_2$	& 17 	&{\bf 16}	& 17	& 20 	& 25		& 32 			&	41	& 52		& 65 	& 80 \\
$D_5$	& 17 	& 10 			&  5	&  2 	&{\bf 1}&  2			&  5	&	10 		& 17	& 26 \\
$D_6$	& 50 	& 41 			& 34	& 29 	& 26 		&{\bf 25}	& 26	&	29 		& 34	& 41 \\
$D_8$	& 53	& 40 			& 29	& 20	& 13 		&  8			&  5	&{\bf 4}&  5	&  8 \\
 	min			& 17 	& 10			&  5	&  2 	&  1		&  2			&	 5 	&  4		&  5 	&  8 \\
 $k^*$		& 2/5	&  5			&  5	&  5	&  5		&  5			&	5/8 &  8		&  8 	&  8
\addpath{(0,7,6)rrrrrrrrrrrrr} 
\addpath{(0,6,6)rrrrrrrrrrrrr} 
\addpath{(0,5,4)rrrrrrrrrrrrr} 
\addpath{(0,4,4)rrrrrrrrrrrrr} 
\addpath{(0,3,4)rrrrrrrrrrrrr} 
\addpath{(0,2,6)rrrrrrrrrrrrr} 
\addpath{(0,1,6)rrrrrrrrrrrrr} 
\addpath{(1,0,6)uuuuuuuu} 
\end{TAB}
}
	\caption{\label{fig_propagation}Propagation of horizontal distances starting from the vertical components $D_{k,{\rm v}}$ taken from the first row of the matrix in Figure \ref{fig_EEDT}~(b). Variable $k^*$ indicates which column contributes the smallest distance $D$.}
\end{figure}

Since we are looking for the minimum distance, the minimum value is determined for each column $j$. This results to a row of distances which is equal to what we already got in Figure \ref{fig_EEDT}~(c), in first row.

Additionally, we can keep track of which column $k$ contributes the minimum distance for a certain column $j_0$. As can be seen in Figure \ref{fig_propagation}, the background pixel at position $j = 1$ has a minimum distance to the object point lying in column $k=2$ and the same distance to the point lying in column $k=5$. The background points in columns $2 \le j\le 6$ are closest to the object point from column $k=5$. For $j=7$, again two object points share by chance the same distance to the background point a.s.o. The object pixel from column $k=6$ is not taken into account for this row because already its vertical component $D_{6,{\rm v}} = 25$ is to high.
		%
\subsubsection{Method by Meijster, Roerdink, and Hesselink }
\begin{figure}
	\hfil \includegraphics[scale=0.8]{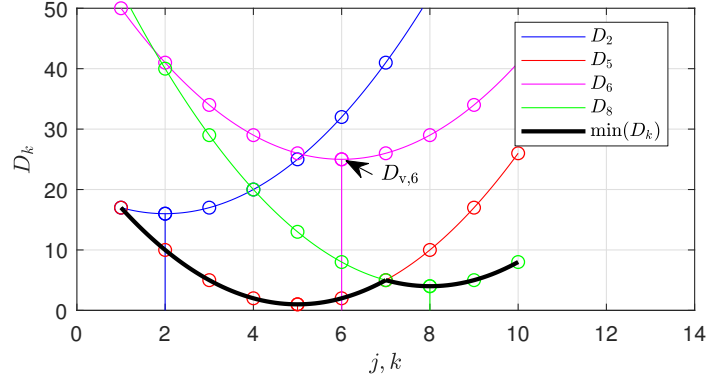}
	\caption{\label{fig_parabola}Graphical visualization of distances $D_k$ from Figure \ref{fig_propagation}. The vertical distances $D_{k,{\rm v}}$ are located at the parabola vertices. Only $\min\limits_k(D_k)$ (lower envelop) is of interest.}
\end{figure}

Looking at Figure \ref{fig_propagation}, it becomes clear that the values in the rows for $D_k$ follow a quadratic equation 
	\begin{align} \label{eq_parabola}
		D_k = D_{k,{\rm v}} + (j -k)^2
	\end{align}
with $j$ being the independent variable. Hence, the distance $D_k=f(j)$ can be visualized as parabolas, see \Figu{fig_parabola}.
When looking for minimal distances $D(i_0, j_0) = \min\limits_k(D_k)$, it becomes clear that only the lower envelop of all parabolas is of interest and object points which do not contribute to this envelop can be ignored in the distance computations.

It also can deduced from Figure \ref{fig_parabola} that the intersection points of contributing parabolas define the range in which a parabola (i.e.\ the according object point) is the closest one. 
This idea has been proposed by Meijster et al \cite{Mei00}. Working source code has been provided by Felsenszwalb and Hutterlocher \cite{Fel12} while it is unclear whether they had been aware of Meijsters work.

So, the key point is to determine the position of the intersection of two parabolas. Let us assume that there are two parabolas with vertices at the columns $k$ and $l$. Since they have a single intersection, the position $j_s$ of this intersection can be determined using (\ref{eq_parabola}) with following equation:
	\begin{align} \label{eq_intersection}
		D_{k,{\rm v}} + (j_s -k)^2 &= D_{l,{\rm v}} + (j_s -l)^2 \nonumber\\
		 (j_s -k)^2 - (j_s -l)^2 &= D_{l,{\rm v}} - D_{k,{\rm v}}  \nonumber\\
		 j_s^2 -2kj_s + k^2 - j_s^2 + 2lj_s - l^2 &= D_{l,{\rm v}} - D_{k,{\rm v}}  \nonumber\\
		2lj_s  -2kj_s  &= D_{l,{\rm v}} - D_{k,{\rm v}} - k^2  + l^2 \nonumber\\
		j_s \cdot 2(l -k)  &= D_{l,{\rm v}} - D_{k,{\rm v}} - k^2  + l^2 \nonumber\\
		j_s &= \frac{ D_{l,{\rm v}} - D_{k,{\rm v}} - k^2  + l^2 }{ 2(l-k) } 
	\end{align}

Using the example in Figure \ref{fig_parabola}, an algorithm is explained in the following, which can significantly reduce the complexity.
Two vectors are needed for storing the required information of the lower envelop. The first vector $ks$ will contain the positions $k\in \{j_n\}$ of all parabolas which finally contribute to the lower envelop and the second vector $js$ will contain the positions $j$ from which a parabola starts to contribute to the lower envelop. Theses positions can be derived from the intersections $j_s$ rounded to the next larger integer value.

The vectors have to be initialized with the parameters of the first, possibly virtual, parabola: $ks(1)=1$ and $js(1)=-\infty$. The huge negative value serves as stopping criterion for the algorithm as will be explained below. The corresponding source code is shown in \Listing{lst_eedtFast} in lines 6 - 12.

The position of the first available (blue) parabola is $k=2$ and the intersection $j_s$ with the virtual parabola is per definition larger than $-\infty$. The information of the new parabola is stored: $ks(2)=2$ and $js(2)=\max(1, j_s)$. It has to be ensured that the lower border starts not before $j=1$.

The position of the next available (red) parabola is $k=5$ and the intersection with the previous is exactly at position $j_s=1$. Since $j_s <= js(2)$ holds, the blue parabola is hidden by the new one and does not contribute anymore. Its parameters can be overwritten: $ks(2) = 5$, $js(2) = 1$. This check has now to be done with all previous parabolas (lines 21-26). As soon as the intersection is $j_s > js(idx)$, the previous parabola contributes to the envelop and the information of the current parabola is appended to the vectors $ks$ and $js$.
The third (magenta) parabola of the example in Figure \ref{fig_parabola} intersect with the previously stored (red) parabola at a position beyond $L$ and thus can be ignored. 
The fourth and last (green) parabola has an intersection with the red one at $j_s=7$. This is larger than the previously stored value of $js(2) = 1$, so it does not hide the previous parabola and the vectors must be complemented with the parameters of the green parabola: $ks(3) = 8$, $js(3) = 7$ (lines 27 - 31). There is no further parabola available and the intersections vector can be closed with $js(1) = 1$ and $js(idx+1) = L+1$ defining the maximum range. The vectors now read as $js=(1, 1, 7, 11)$ and $ks=(1, 5, 8)$ giving the information that parabola at $k=1$ is either not existing or not contributing ($1\le j < 1$~!), parabola $k=5$ defines the distances for the positions $1\le j <7$ and parabola $k=8$ is responsible for the distances at $7\le j <11$. Using equation (\ref{eq_parabola}), the squared Euclidean distance can be calculated for all positions $j$ (lines 37-42). This procedure is performed for each row of the distance matrix.

In total, each row is scanned twice. The first scan identifies the contributing object points (parabolas) and the second scan applies the collected information and calculates the final distances $D(i:0, j_0)$.
Even though, the first scan uses a double while-loop, its complexity can be considered as being linear ($O(L)$). That means, the entire method has $O(M\cdot L)$ complexity.

    %

	%
\section{Related things}
   %
The method  by Meijster {\it et al}	is not the only possibility for the efficient determination of relevant object points. Maurer {\it et al} proposed a technique that is based on the computation of lines separating the regions which belong to different object pixels \cite{Mau03}, see Figure \ref{fig_propError}. This method also efficiently identifies the contributing object points for each row and reaches $O(M\cdot L)$ complexity.

\appendix

\section{Source Code}
    %
	\noindent\begin{minipage}[t]{.5\textwidth}
\begin{lstlisting}[caption={\raggedright Examples for distance calculations \phantom{WWWWWWWWWWW} },label=lst_distCalc]
N = 6;
disp('Euclidean') % Euclidean distances
A = ((0:N-1).^2 )' * ones(1, N);
B = A';
E = A+B % squared Euclidean distances

disp('Chamfer') % Chamfer distances
C = zeros(N);
diagStep = sqrt(2);
%diagStep = 4./3;
for i = 1:N
  for j = 1:N
    C(i,j) = j-1 + i-1;
    if i > 1 && j > 1
      C(i,j) = C(i-1,j-1) + diagStep;
    end
  end
end
Cr = round( 10* C.^2) / 10 % rounded squared Chamfer distances
D = sqrt(E) - sqrt(Cr) % difference between Euclidean and Chamfer

sum( abs(D(:))) / (N*N) % average error
max(abs(D(:))) % max absolute error
\end{lstlisting}
\end{minipage}
	%
	\hfil
\noindent\begin{minipage}[t]{.5\textwidth}
\begin{lstlisting}[caption={\raggedright Example of sequential city-block distance transformation \phantom{WWWWWWWWWWW} },label=lst_cityblockDTsequ]
L = 10; %number of columns
M = 9; % number of rows
maxDist = L + M; % sufficient large distance
distMat = ones(M,L) * maxDist; % initilize 

% example coordinates of object pixels
x = [ 2, 5, 6, 8, 8, 8]; % horizontal position
y = [ 5, 2, 6, 3, 4, 7]; % vertical position

for n = 1: length (x)
  distMat(y(n), x(n)) = 0; % set object pixels
end

% scan top line from left to right
i = 1;
for j = 2:L % ignore left border
    if distMat(i,j) > distMat( i, j-1) + 1
       distMat(i,j) = distMat( i, j-1) + 1;
    end
end
% scan from top-left to bottom-right
for i = 2:M % leave top  border
  for j = 2:L % leave left  border
    if distMat(i,j) > distMat( i-1, j) + 1
        distMat(i,j) = distMat( i-1, j) + 1;
    end
    if distMat(i,j) > distMat( i, j-1) + 1
        distMat(i,j) = distMat( i, j-1) + 1;
    end
  end
end

% scan bottom line from right to left
for j = L-1:-1:1 % ignore left border
    if distMat(M,j) > distMat( M, j+1) + 1
       distMat(M,j) = distMat( M, j+1) + 1;
    end
end
% scan right column from bottom to top
for i = M-1:-1:1 % ignore left border
    if distMat(i,L) > distMat( i+1, L) + 1
       distMat(i,L) = distMat( i+1, L) + 1;
    end
end
% scan from bottom-right to top-left
for i = M-1:-1:1 % ignore bottom border
  for j = L-1:-1:1 % ignore right border
    % look backwards
    if distMat(i,j) > distMat( i+1, j) + 1
       distMat(i,j) = distMat( i+1, j) + 1; 
    end
    if distMat(i,j) > distMat( i, j+1) + 1
       distMat(i,j) = distMat( i, j+1) + 1; 
    end
  end
end
distMat % outputs the resulting distances
\end{lstlisting}
\end{minipage}
	\vfil
	~\\

	%
\noindent\begin{minipage}[t]{.5\textwidth}
\begin{lstlisting}[caption={\raggedright Example of parallel city-block distance transformation \phantom{WWWWWWWWWWW} },label=lst_cityblockDTpara]
L = 10; %number of columns
M = 9; % number of rows
maxDist = L + M; % sufficient large distance
distMat = ones(M,L) * maxDist; % initilize 

% coordinates of object pixels
x = [ 2, 5, 6, 8, 8, 8]; % horizontal position
y = [ 5, 2, 6, 3, 4, 7]; % vertical position

for n = 1: length (x)
  distMat(y(n), x(n)) = 0; % set object pixels
end

% scan from top to bottom and back
for j = 1:L % for all columns
  % downward
  for i = 2:M % leave top border
    if distMat(i,j) > distMat( i-1, j) + 1
        distMat(i,j) = distMat( i-1, j) + 1;
    end
  end
  % upward
  for i = M-1:-1:1 % leave bottom border
    if distMat(i,j) > distMat( i+1, j) + 1
        distMat(i,j) = distMat( i+1, j) + 1;
    end
  end
end

% scan from left to right and back
for i = 1:M % for all rows
  for j = 2:L% ignore left border
    % to the right
    if distMat(i,j) > distMat( i, j-1) + 1
       distMat(i,j) = distMat( i, j-1) + 1; 
    end
  end
  for j = L-1:-1:1% ignore left border
    % to the left
    if distMat(i,j) > distMat( i, j+1) + 1
       distMat(i,j) = distMat( i, j+1) + 1; 
    end
  end
end

distMat % outputs the resulting distances
\end{lstlisting}
\end{minipage}
	\hfil
\noindent\begin{minipage}[t]{.5\textwidth}
\begin{lstlisting}[caption={\raggedright Example of approximate Eucidean distance transformation (part 1) \phantom{WWWWWWWWWWW} },label=lst_DanielssonDT]
L = 10; %number of columns
M = 9; % number of rows
maxDist = L*L + M*M; % sufficient large distance
distMat = ones(M,L) * maxDist; % distance matrix
 % initilize relative positions
distMatX = ones(M,L) * maxDist;
distMatY = ones(M,L) * maxDist; 
% matrix for pre-calculated distances
preCalc = -ones( max(M, L)); 

% coordinates of object pixels
x = [ 2, 5, 6, 8, 8, 8]; % horizontal position
y = [ 5, 2, 6, 3, 4, 7]; % vertical position

% set object pixels
for n = 1: length (x)
  distMat(y(n), x(n)) = 0; 
  distMatY(y(n), x(n)) = 0; 
  distMatX(y(n), x(n)) = 0; 
end

% first scan 
:
% secon scan 
:

---------------------------
function [dist] = getDist( dy, dx, preCalc)
	% '+1' because of matlab indexing
  if preCalc(dy+1, dx+1) < 0 
      dist = dy * dy + dx * dx;
      preCalc(dy+1, dx+1) = dist;
      preCalc(dx+1, dy+1) = dist; % symmetric
  else % use pre-calculation
    dist = preCalc(dy+1, dx+1);
  end
end
\end{lstlisting}
\end{minipage}
		\hfil
\noindent\begin{minipage}[t]{.5\textwidth}
\begin{lstlisting}[caption={\raggedright Example of approximate Eucidean distance transformation (part 2) \phantom{WWWWWWWWWWW}},label=lst_DanielssonDT2]
% first scan
for i = 2:M % exclude top border
  for j = 1:L % 
    yr = distMatY(i-1,j); % top neighbour
    if yr < maxDist % something to propagate
      xr = distMatX(i-1,j);
      dist = getDist( yr+1, xr); % new distance
      if distMat(i,j) > dist
        distMat(i,j) = dist; % replace
        distMatY(i,j) = yr+1;
        distMatX(i,j) = xr;
      end
    end
  end
  for j = 2:L % exclude left border
      yr = distMatY(i,j-1); % left neighbour
      if yr < maxDist
        xr = distMatX(i,j-1);
        dist = getDist( yr, xr+1);
        if distMat(i,j) > dist
          distMat(i,j) = dist;
          distMatY(i,j) = yr;
          distMatX(i,j) = xr+1;
        end
      end
      yr = distMatY(i-1,j-1); % top-left neighbour
      if yr < maxDist
        xr = distMatX(i-1,j-1);
        dist = getDist( yr+1, xr+1);
        if distMat(i,j) > dist
          distMat(i,j) = dist;
          distMatY(i,j) = yr+1;
          distMatX(i,j) = xr+1;
        end
      end
  end
  for j = L-1:-1:1 % backward
      yr = distMatY(i-1,j+1); % top-right neighbour
      if yr < maxDist
        xr = distMatX(i-1,j+1);
        dist = getDist( yr+1, xr+1);
        if distMat(i,j) > dist
          distMat(i,j) = dist;
          distMatY(i,j) = yr+1;
          distMatX(i,j) = xr+1;
        end
      end
      yr = distMatY(i,j+1);
      if yr < maxDist
        xr = distMatX(i,j+1); % right neighbour
        dist = getDist( yr, xr+1);
        if distMat(i,j) > dist
          distMat(i,j) = dist;
          distMatY(i,j) = yr;
          distMatX(i,j) = xr+1;
        end
      end
  end
end
\end{lstlisting}
\end{minipage}
		\hfil
\noindent\begin{minipage}[t]{.5\textwidth}
\begin{lstlisting}[caption={\raggedright Example of approximate Eucidean distance transformation (part 3) \phantom{WWWWWWWWWWW}},label=lst_DanielssonDT3]
% second scan 
for i = M-1:-1:1 % ignore bottom border
  for j = 1:L 
    yr = distMatY(i+1,j);
    if yr < maxDist
      xr = distMatX(i+1,j);
      dist = getDist( yr+1, xr);
      if distMat(i,j) > dist
        distMat(i,j) = dist;
        distMatY(i,j) = yr+1;
        distMatX(i,j) = xr;
      end
    end
  end
  for j = 2:L % ignore left border
    yr = distMatY(i,j-1);
    if yr < maxDist
      xr = distMatX(i,j-1);
      dist = getDist( yr, xr+1);
      if distMat(i,j) > dist
        distMat(i,j) = dist;
        distMatY(i,j) = yr;
        distMatX(i,j) = xr+1;
      end
    end
    yr = distMatY(i+1,j-1);
    if yr < maxDist
      xr = distMatX(i+1,j-1);
      dist = getDist( yr+1, xr+1);
      if distMat(i,j) > dist
        distMat(i,j) = dist;
        distMatY(i,j) = yr+1;
        distMatX(i,j) = xr+1;
      end
    end
  end
  for j = L-1:-1:1 % ignore right border
    yr = distMatY(i,j+1);
    if yr < maxDist
      xr = distMatX(i,j+1);
      dist = getDist( yr, xr+1);
      if distMat(i,j) > dist
        distMat(i,j) = dist;
        distMatY(i,j) = yr;
        distMatX(i,j) = xr+1;
      end
    end
    yr = distMatY(i+1,j+1);
    if yr < maxDist
      xr = distMatX(i+1,j+1);
      dist = getDist( yr+1, xr+1);
      if distMat(i,j) > dist
        distMat(i,j) = dist;
        distMatY(i,j) = yr+1;
        distMatX(i,j) = xr+1;
      end
    end
  end
end
\end{lstlisting}
\end{minipage}
\begin{minipage}[t]{.5\textwidth}
\begin{lstlisting}[caption={\raggedright Simple EEDT algorithm},label={lst_eedtSimple}]
L = 10; %number of columns
M = 9; % number of rows
maxDIst = L*L + M*M; % sufficient large distance
distMat = ones(M,L) * maxDist; % initilize distance matrix

% coordinates of object pixels
x = [ 2, 5, 6, 8, 8, 8]; % horizontal position
y = [ 5, 2, 6, 3, 4, 7]; % vertical position

for n = 1: length (x)
  distMat(y(n), x(n)) = 0; % set object pixels
end

% assign distances column-wise
for j = 1:L % for all columns
  distStep = 1;
  for i = 2:M % propagate distances downwards
    if distMat(i,j) > distMat(i-1,j) + distStep
      distMat(i,j) = distMat(i-1,j) + distStep;
      distStep = distStep + 2;
    else
      distStep = 1;
    end
  end
  distStep = 1;
  for i = M-1:-1:1 % propagate distances upwards 
    if distMat(i,j) > distMat(i+1,j) + distStep
      distMat(i,j) = distMat(i+1,j) + distStep;
      distStep = distStep + 2;
    else
      distStep = 1;
    end
  end
end

% determine distances row-wise
for i = 1:M % all rows
	% copy row of vertical distances
	distMatV = distMat(i,:); 
  for j = 1:L % column positions
		% initialize minimum distance
    distMin = distMatV(j); 
    for k = 1:L % compare to column positions
      % combine vert. with horiz. component
      dist = distMatV(k) + (k-j)*(k-j); 
      if distMin > dist
        distMin = dist; % new minimum distance
      end
    end
    distMat(i,j) = distMin; % assign minimum
  end
end

distMat % outputs the resulting squared distances
\end{lstlisting}
\end{minipage}
	\hfil
\begin{minipage}[t]{.5\textwidth}
\begin{lstlisting}[caption={\raggedright Improved row processing of simple EEDT algorithm \phantom{WWWWWWW} \phantom{WWWWWW} },label={lst_eedtImproved}]
initialization
 :
determination of vertical distances
 :
% determine distances row-wise
for i = 1:M % all rows
	% copy row of vertical distances
	distMatV = distMat(i,:); 
  for j = 1:L % column positions j0
		% initialize minimum distance
    distMin = distMatV(j); 
    for k = j+1:L % compare to column positions forward
      if distMatV(k) < maxDist
        % combine vert. with horiz. component
        distHor = (k-j)*(k-j);
        if distHor >= distMin
          break; % pure horizontal component is longer then current distance
        end
        dist = distMatV(k) + distHor; 
        if distMin > dist
          distMin = dist; % store new minimum distance
        end
      end
    end
    for k = j-1:-1:1 % compare to column positions backward
      if distMatV(k) < maxDist
        % combine vertical with horizontal component
        distHor = (k-j)*(k-j);
        if distHor >= distMin
          break; % pure horizontal component ...
        end
        dist = distMatV(k) + distHor; 
        if distMin > dist
          distMin = dist; % store new minimum distance
        end
      end
    end
    distMat(i,j) = distMin; % assign minimum
  end
end

distMat % outputs the resulting squared distances
\end{lstlisting}
\end{minipage}

		\hfil
\begin{minipage}[t]{.8\textwidth}
\begin{lstlisting}[caption={\raggedright Fast algorithm for EEDT},label={lst_eedtFast}]
initialization
 :
determination of vertical distances
 :
% determine distances row-wise
js = zeros( 1, L+1); % stores intersection positions
ks = zeros( 1, L); %  positions of contributors
for i = 1:M % all rows
  distMatV = distMat( i,:); %copy row of distances
  idx = 1;
  js(1) = -maxDist; % serves as stopping point
  ks(1) = 1; % assume first (possibly dummy) contributor
  m = 1; % take parabola at first column (if any) as given
  while m < L % look for next contributor
    m = m + 1;
    if distMatV(m) < maxDist
      mm = m * m;
      % compute intersection with previous contributor 
      k = ks(idx); % position of previous
      j = ceil( (distMatV(m) - distMatV(k) - k*k + mm) / ( 2*(m-k) ));
      while j <= js(idx) % new parabola hides previous
        idx = idx - 1; % go one element back
        k = ks(idx); % position of previous
				% compute intersection with previous contributor 
        j = ceil( (distMatV(m) - distMatV(k) - k*k + mm) / ( 2*(m-k) ));
      end
      if j <= L % make sure that new parabola contributes inside matrix
        idx = idx + 1; % store parameters in next elements
        js(idx) = max(1,j); % save new intersection, keep it inside range
        ks(idx) = m; % save column of next contributor
      end
    end
  end
  js(1) = 1; % left border of contribution
  js(idx+1) = L+1; % right border of contribution
  % now apply collected information (lower envelop)
  for n = 1:idx % for all stored contributors
    k = ks(n); % column of contributor
    for j = js(n) : js(n+1)-1 % get region of contribution
      distMat(i,j) = distMatV(k) + (j-k) * (j-k); % assign  distance
    end
  end
end

distMat % outputs the resulting squared distances
\end{lstlisting}
\end{minipage}

	%
    %

	%
	%
\bibliography{literature} 

\begin{thebibliography}{Mau03}
\providecommand{\url}[1]{\texttt{#1}}
\providecommand{\urlprefix}{URL }
\input{babelbst.tex}
\newcommand{\Capitalize}[1]{\uppercase{#1}}
\newcommand{\capitalize}[1]{\expandafter\Capitalize#1}
\providecommand{\eprint}[2][]{\url{#2}}

\bibitem[Bai05]{Bai04}
Bailey D.G.: An efficient {Euclidean} distance transform. R.~Klette;
  J.~{\v{Z}}uni{\'{c}} (\capitalize\bbleditors{}), \emph{Combinatorial Image
  Analysis}, Springer, Berlin, Heidelberg, 2005, 394--408

\bibitem[Bor84]{Bor84}
Borgefors G.: Distance transformations in arbitrary dimensions. \emph{Computer
  Vision, Graphics, and Image Processing}, \bblvol{}~27, 1984, 321 -- 345

\bibitem[Dan80]{Dan80}
Danielsson P.E.: Euclidean distance mapping. \emph{Computer Graphics and Image
  Processing}, \bblvol{}~14, \bblno{}~3, 1980, 227--248,
  \urlprefix\url{https://www.sciencedirect.com/science/article/pii/0146664X80900544}

\bibitem[Fab08]{Fab08}
Fabbri R.; Costa L.D.F.; Torelli J.C.; Bruno O.M.: {2D Euclidean} distance
  transform algorithms: A comparative survey. \emph{ACM Comput. Surv.},
  \bblvol{}~40, \bblno{}~1, \bblfeb{} 2008,
  \urlprefix\url{https://doi.org/10.1145/1322432.1322434}

\bibitem[Fel12]{Fel12}
Felzenszwalb P.; Huttenlocher D.: Distance transforms of sampled functions.
  \emph{Theory of Computing}, \bblvol{}~8, September 2012, 415 -- 428

\bibitem[Gre07]{Grevera2007}
Grevera G.J.: Distance transform algorithms and their implementation and
  evaluation. \emph{Deformable Models: Biomedical and Clinical Applications},
  Springer New York, New York, NY, 2007, 33--60,
  \urlprefix\url{https://doi.org/10.1007/978-0-387-68413-0_2}

\bibitem[Mau03]{Mau03}
Maurer C.; Qi R.; Raghavan V.: A linear time algorithm for computing exact
  euclidean distance transforms of binary images in arbitrary dimensions.
  \emph{IEEE Transactions on Pattern Analysis and Machine Intelligence},
  \bblvol{}~25, \bblno{}~2, 2003, 265--270

\bibitem[Mei00]{Mei00}
Meijster A.; Roerdink J.; Hesselink W.: A general algorithm for computing
  distance transforms in linear time. J.~Goutsias; L.~Vincent; D.~Bloomberg
  (\capitalize\bbleditors{}), \emph{Mathematical morphology and its
  applications to image and signal processing}, Computational imaging and
  vision, University of Groningen, Johann Bernoulli Institute for Mathematics
  and Computer Science, 2000, 331--340, relation:
  http://www.rug.nl/informatica/onderzoek/bernoulli; 5th International
  Symposium on Mathematical Morphology (ISMM); Conference date: 26-06-2000
  Through 29-06-2000

\bibitem[Ros68]{Ros68}
Rosenfeld A.; Pfaltz J.: Distance functions on digital pictures. \emph{Pattern
  Recognition}, \bblvol{}~1, 1968, 33 -- 61

\end{thebibliography}
\bibliographystyle{alphaTSnew}



	%
\end{document}